\documentclass[10pt,twocolumn,letterpaper]{article}

\usepackage[pagenumbers]{wacv} %

\usepackage{graphicx}
\usepackage{amsmath}
\usepackage{amssymb}
\usepackage{booktabs}
\usepackage{pgfplots}
\pgfplotsset{width=\textwidth,compat=1.9, height=100pt} %
\usepackage{adjustbox}
\usepackage{tikz-3dplot}
\usepackage[acronym]{glossaries}
\usepackage{calc}
\usepackage{tikz}
\usepackage[percent]{overpic}
\usetikzlibrary{backgrounds}
\usetikzlibrary{fit}
\usetikzlibrary{positioning} %
\usetikzlibrary{calc}
\usetikzlibrary{patterns} %
\usetikzlibrary{quotes} %
\usetikzlibrary{calc}
\usetikzlibrary{intersections}

\usepackage[pagebackref,breaklinks,colorlinks]{hyperref}

\usepackage[capitalize]{cleveref}
\crefname{section}{Sec.}{Secs.}
\Crefname{section}{Section}{Sections}
\Crefname{table}{Table}{Tables}
\crefname{table}{Tab.}{Tabs.}

\def\wacvPaperID{****} %
\def\confName{WACV}
\def\confYear{2025}

\begin{document}

\newcommand{\thoughts}[1]{\textcolor{red}{#1}}
\newacronym{cd}{CD}{Center Distance}
\newacronym{iou}{IoU}{Intersection over Union}
\newacronym{aabb}{AABB}{Axis-Aligned Bounding Box}
\newacronym{roi}{ROI}{Region of Interest}
\newacronym{ev}{EV}{Extra Volume}
\newacronym{esf}{ESF}{Encompassment Scaling Factor}
\newacronym{ct}{CT}{Computed Tomography}
\newacronym{mri}{MRI}{Magnetic Resonance Imaging}
\newacronym{as}{ANS}{Anatomical Structure}
\newacronym{sdf}{SDF}{Signed Distance Function}
\newacronym{mlp}{MLP}{Multilayer Perceptron}
\newacronym{icp}{ICP}{Iterative Closest Point}

\title{LOOC: Localizing Organs using Occupancy Networks \\and Body Surface Depth Images}

\author{Pit Henrich and $^*$Franziska Mathis-Ullrich\\
Department of Artificial Intelligence in Biomedical Engineering, Friedrich-Alexander-University\\Erlangen-Nürnberg, 91052 Erlangen\\
{\tt\small \{pit.henrich,franziska.mathis-ullrich\}@fau.de}\\
\small $^*$Corresponding Author
}

\twocolumn[{%
\renewcommand\twocolumn[1][]{#1}%
\maketitle
\begin{center}
    \centering
    \captionsetup{type=figure}
    \begin{tikzpicture}
        \node[anchor=south west,inner sep=0] (image) at (0,0) {
        \centering
        \includegraphics[width=0.90\textwidth]{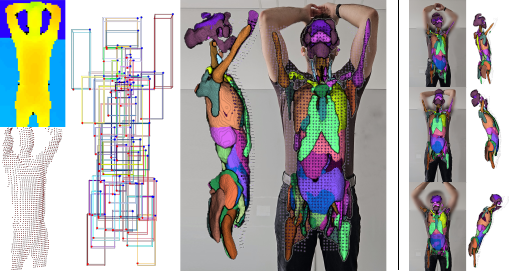}};
        \begin{scope}[
            x={(image.south east)},
            y={(image.north west)}
        ]
            \node [white, font=\bfseries] at (0.112,0.57) {(a)};
            \node [black, font=\bfseries] at (0.112,0.07) {(b)};
            \node [black, font=\bfseries] at (0.325,0.07) {(c)};
            \node [black, font=\bfseries] at (0.725,0.07) {(d)};
            \node [black, font=\bfseries] at (0.950,0.07) {(e)};
        \end{scope}
    \end{tikzpicture}
    \captionof{figure}{
        A real-world depth image (a) is converted to a point cloud (b).
        Our occupancy network, conditioned on the point cloud, estimates the bounding boxes of 67 anatomical structures (c).
        It also generates a patient specific 3D anatomical atlas (d).
        As shown in (e), changes in the patient's body pose are reflected in the output.
    }
    \label{fig:overview}
\end{center}%
}]
\maketitle

\begin{abstract}
    We introduce a novel approach for the precise localization of 67 anatomical structures from single depth images captured from the exterior of the human body.
    Our method uses a multi-class occupancy network, trained using segmented CT scans augmented with body-pose changes, and incorporates a specialized sampling strategy to handle densely packed internal organs.
    Our contributions include the application of occupancy networks for occluded structure localization, a robust method for estimating anatomical positions from depth images, and the creation of detailed, individualized 3D anatomical atlases.
    We outperform localization using template matching and provide qualitative real-world reconstructions.
    This method promises improvements in automated medical imaging and diagnostic procedures by offering accurate, non-invasive localization of critical anatomical structures.
\end{abstract}

\section{Introduction}

\newdimen\OccupancyNetworkX
\newdimen\OccupancyPointCloudY

\newcommand{\graphicnodeapp}[3]{

    \node (#2) [align=center, #1] {
        \shortstack{
            \includegraphics[height=5cm]{#3} \\
            \parbox{5cm}{}
        }
    };
}

\newcommand{\textnodeapp}[2]{

    \node (#2) [draw, rectangle, align=center, #1] {
        \shortstack{
            \parbox{5cm}{}
        }
    };
}

\begin{figure}
    \centering
    \resizebox{0.95\columnwidth}{!}{
            \begin{tikzpicture}[node distance=0.3cm, auto]
            \begin{scope}[local bounding box=firstBox]

                \node (A) [align=center] {\includegraphics[height=5cm]{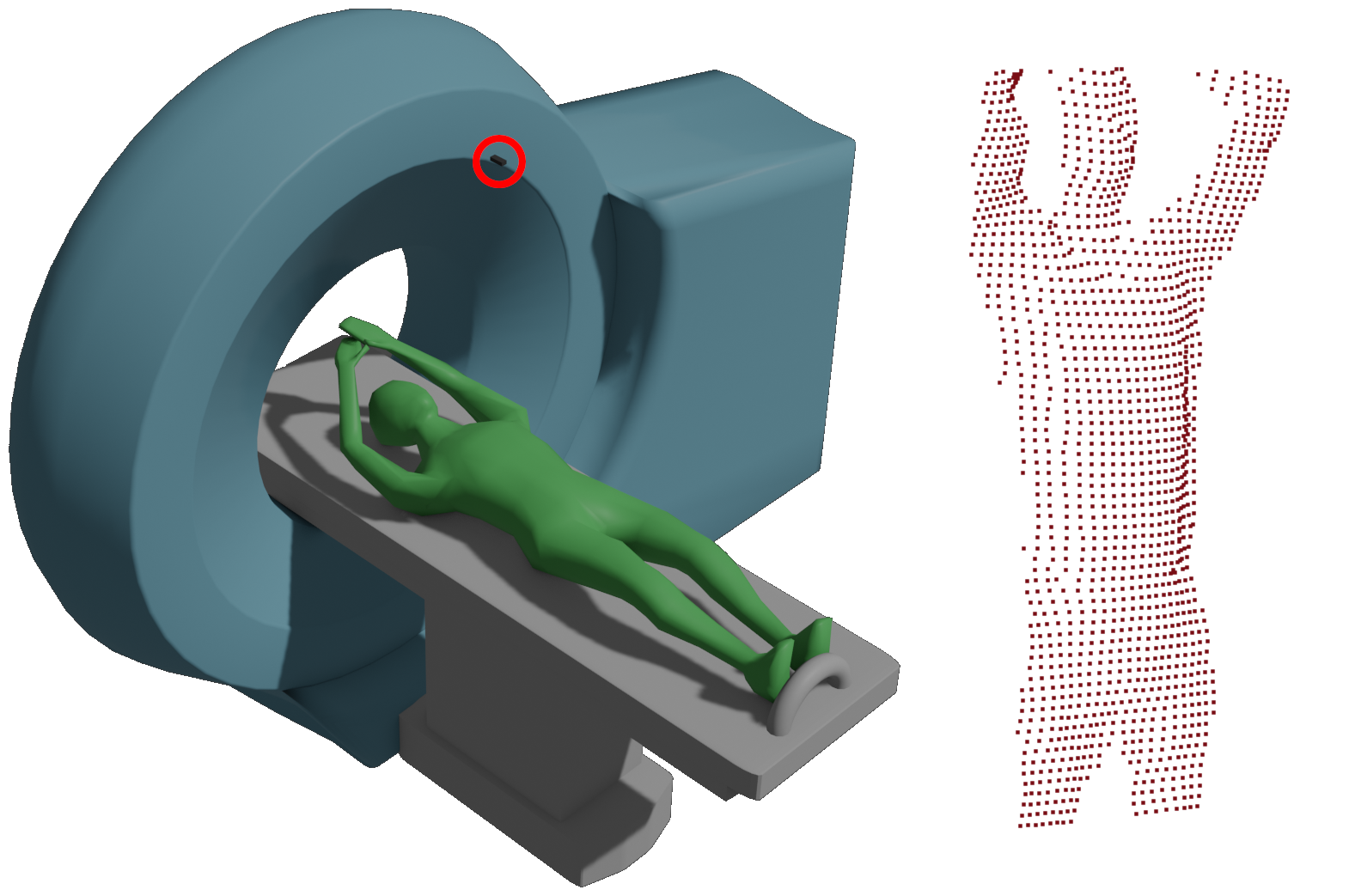}};
                \graphicnodeapp{below=of A}{B}{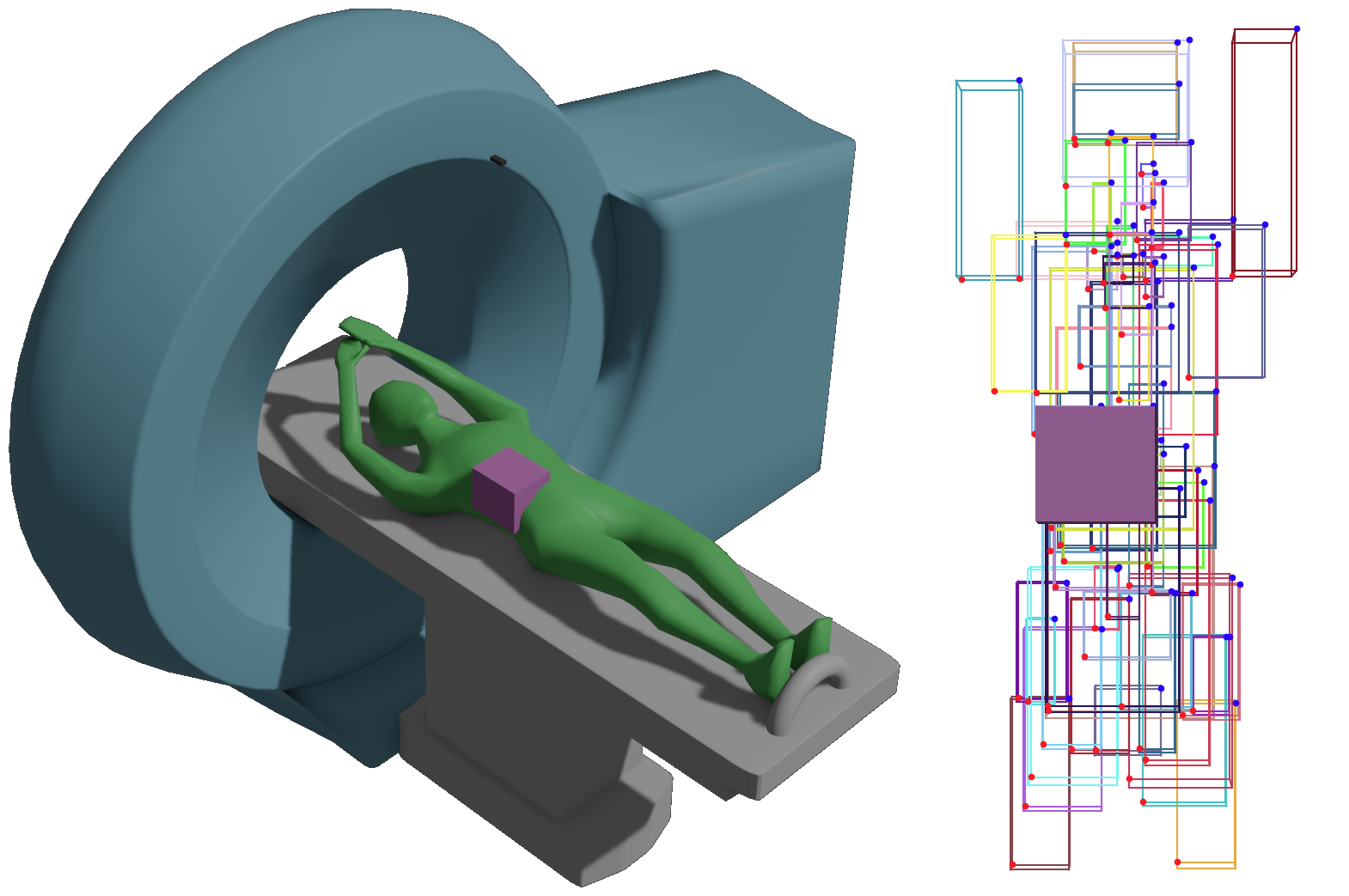}
                \graphicnodeapp{below=of B}{C}{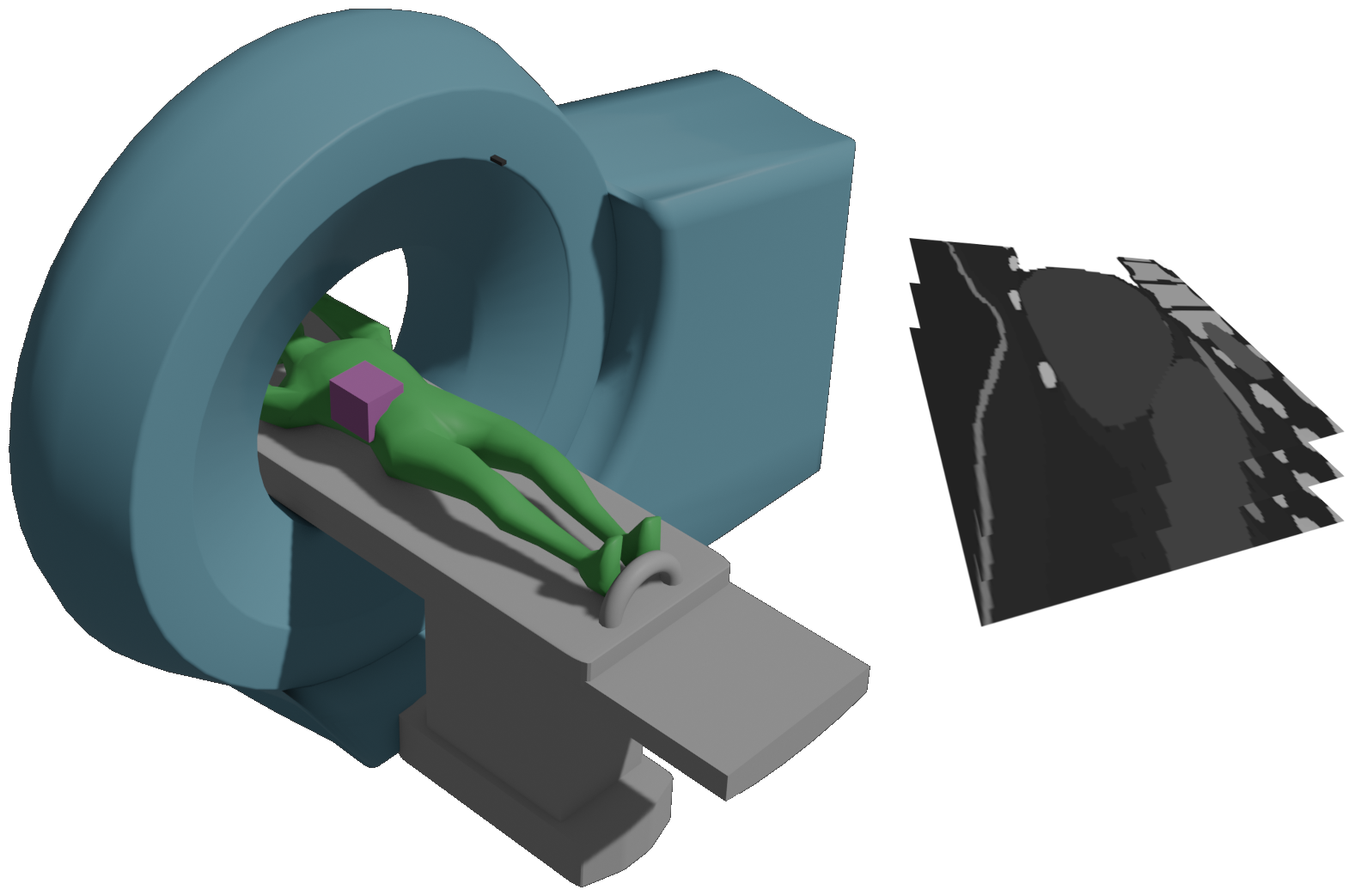}
        
                \draw[thick,->] (A) -- (B);
                \draw[thick,->] (B) -- (C);

                \node (Depth Sensor) at (-4,4) {Depth Sensor};
                \node (Medical Scanner) at (-1,3) {Medical Scanner};
                \node (Patient) at (-4,-2) {Patient};
                \draw[thick, ->, color=magenta] (Depth Sensor) |- (-1.2, 1.6);
                \draw[thick, ->, color=magenta] (Patient) |- (-1, -0.2);
                \draw[thick, ->, color=magenta] (Medical Scanner.south) -| (0,1);

                \node (ROI) at (-3.5,-7.5) {ROI (AABB)};
                \draw[thick, ->, color=magenta] (ROI) |- (-1.0, -5.8);
                \draw[thick, ->, color=magenta] (ROI) |- (2.1, -5.6);

                \node (Scan Region) at (-3.0,-13.5) {Scan Region};
                \draw[thick, ->, color=magenta] (Scan Region) |- (-1.8, -11);

                \node (Sensor Point Cloud) at (2.5,2.5) {Sensor Point Cloud};
                \node (AABB of ANS) at (2.5,-2.8) {AABB of ANS};
                \node (Image Stack) at (2.5,-9.5) {Image Stack};

            \end{scope}
            \end{tikzpicture}
        
      }
    \caption{A \textit{Patient} lying on the insertion table of a \textit{Medical Scanner}. The \textit{Sensor Point Cloud} from the fixed \textit{Depth Sensor} is used to automatically estimate Axis Aligned Bounding Boxes (AABBs) for $67$ Anatomical Structures (ANSs). The AABB of the wanted ANS is considered the ROI.
    The ROI is used as the \textit{Scan Region}, which is imaged by the \textit{Medical Scanner} to produce a volumetric \textit{Image Stack}.}
    \label{fig:application}
\end{figure}

Multiple medical imaging tasks require an initial location estimate of \glspl{as} of interest.
For example, an ultrasound probe needs to be positioned such that it visualizes a structure of interest or to begin visual servoing~\cite{Moradi2022, jiang2023robotic}.
For X-ray, \gls{ct}, and \gls{mri}, a \gls{roi} needs to be selected on the patient's body.
Accurately estimating the position of an \gls{as} can minimize scan times or radiation doses~\cite{SHETTY2023107383}. 
However, estimating \glspl{as} is challenging due to the large variability in human body shapes.
In practice, medical staff rely on anatomical knowledge and experience to estimate the locations of \glspl{as} from outside the body.
In computer-assisted interventions, automatic localization of \glspl{as} within the human body enables various downstream tasks, potentially even allowing non-specialists to perform them.
In this work, we estimate both the position and scale of human \glspl{as} using a single depth image.
A radiological example application is shown in \Cref{fig:application}.

\subsection{Related Work}
Medical automation methods, such as robotic ultrasound, often rely on registration methods for path planning~\cite{annurev-control}.
For this, preoperative \gls{ct} or \gls{mri} data is registered onto the patient body using RGB-D data~\cite{modrzejewski2018soft} of the patient skin or the ultrasound image itself for internal structure registration.
Both methods require a preoperatively acquired model.

Comte et al.~\cite{comte20233d} show that the deformations of the spine can be estimated from depth images of the back.
Besides possible applications in task automation, this method also allows the automatic detection of scoliosis with an accuracy of $89\%$.

BOSS~\cite{SHETTY2023107383} is a closed-source industrial solution using statistical shape models.
This method is able to locate 12 anatomical structures based on patient metadata or skin surface models.

Rather than registering the patient to a prior scan or template, an atlas based on the point clouds obtained from RGB-D data can be created.
The inference of 3D objects based on point clouds is a growing field.
Directly generating surfaces meshes is challenging because of the need for regularizers that ensure good geometry and topology.
Convolutional methods~\cite{choy20163d,girdhar2016learning} that generate voxel-based objects suffer from a memory-resolution tradeoff, rendering them unsuitable to the generation of objects with large structures and fine details.
DeepSDF~\cite{park2019deepsdf} and Occupancy Networks~\cite{mescheder2019occupancy} represent objects using continuous implicit functions, approximated by neural networks.
This enables them to construct both large structures and fine details, without the memory limitations of voxel-based methods.
Because occupancy networks can use a cross-entropy loss, they are more easily extended to allow the reconstruction of objects with multiple parts.
They represent an object by approximating a function which assigns any point in space an \textit{inside} or \textit{outside} value.
The boundaries of inside-value-clusters represent a 3D object.
Occupancy labels are values assigned to points in 3D space, indicating whether a point lies inside an object ($\text{value} > 0$) or outside ($\text{value} = 0$).
An occupancy network is trained using these labeled point clouds, which we refer to as occupancy samples.
The network needs to predict the labels of each point during supervised learning.
The labeling can be conditioned using a secondary input, such as a depth image or a point cloud.

Henrich et al.~\cite{henrich2024registered} demonstrate that single view point clouds can be used to reconstruct deformed objects consisting of multiple parts using a multi-class occupancy function.
They propose the use of occupancy networks~\cite{mescheder2019occupancy} in combination with PointNet++~\cite{qi2017pointnet++} to perform a task similar to 3D object registration.
For this, an observation of the deformable object is used to infer a 3D object that best explains this observation.
Crucially, the object whose surface is visible is reconstructed.
To achieve this, a prior 3D object is deformed to produce both ground-truth occupancy samples and sensor-based point clouds.
By being presented many combinations of occupancy samples and sensor point clouds, the system learns to label the occupancy samples through supervised learning.
The trained network is then queried with a new sensor point cloud and equidistant queries to produce a 3D object.

To train a multi-class occupancy network, 3D objects are needed to generate training data.
Several approaches exist for creating synthetic human bodies of various shapes and sizes~\cite{SMPL:2015,SMPL-X:2019,STAR:2020}.
These approaches have been applied by Keller et al.~\cite{Keller:CVPR:2022} to estimate the skeleton shape from an outside image in OSSO.
However, many medical applications focus on other \glspl{as} of interest.
Therefore, Keller et al.~\cite{Keller_2024_CVPR} present HIT as a method to estimate two additional types body tissue. In addition to bone structures, they estimate fat under the skin and lean tissue.
The body fat to lean tissue ratio can provide important medical information, for example to predict the risk for cardiovascular diseases~\cite{grundy2004obesity}.
Still, their approach only provides three classes of tissue, all of which span the whole body.
This renders their approach unsuitable for specific \gls{as} localization tasks.

Instead of generating synthetic human models, training data can be generated from 3D segmentation masks.
Jaus et al. presented the Atlas Dataset~\cite{jaus2023towards}.
Using nnU-Net~\cite{isensee2021nnu}, they produced segmentation masks of 533 human bodies from the AutoPET dataset~\cite{gatidis2022whole}.
These segmentation masks provide structural information for 142 \glspl{as}.

\subsection{Contribution}
We apply occupancy networks to the task of localizing occluded structures.
For this, we present a revised SortSample~\cite{henrich2024registered} that allows the learning of tightly packed objects.
We provide a method to estimate the locations of 67 \glspl{as} from a single depth image, thereby notably surpassing closed-source commercial systems~\cite{SHETTY2023107383}.
We show its advantages over template matching and provide real-world reconstruction examples.
Further, our method produces an estimated 3D anatomical atlas specific to individual patients.

\section{Method}
    We propose a proxy task to solve the problem of locating \glspl{as}.
    We approximate a patient specific \gls{as} atlas based on a single-view point cloud obtained from a depth image.
    This atlas is used to estimate the \glspl{roi} of each \gls{as}.
    For this, each \gls{aabb} is considered the \gls{roi} of the corresponding \gls{as}.

    \subsection{Preliminaries on Occupancy Networks}
        We use occupancy learning in combination with \gls{sdf} learning.
        This combination has been shown to be beneficial to 3D reconstruction accuracy~\cite{lamb2022deepjoin, henrich2024registered}.
        But unlike previous work on 3D reconstruction~\cite{mescheder2019occupancy, park2019deepsdf, henrich2024registered} we do not reconstruct the surface visible to the camera.
        Although the skin surface can be reconstructed, privacy concerns must be taken into account.
        If the training data instills real patient skin reconstructions, the occupancy network may generate visually identifiable individuals.
        To ensure privacy, we do not reconstruct skin.
        This makes the use of autodecoder architectures~\cite{park2019deepsdf} unsuitable.
        Autodecoders perform an inference time optimization to compute a latent representation.
        The objective being a latent vector $\ell$ that ensures the decoder assigns all sensor point cloud points a value close to $0$.
        The latent can be approximated by evaluating $\textit{argmin}_\ell \sum_{i} \left| f_{\text{decoder}}(p_i; \ell) \right|$, where $f_{\text{decoder}}(p_i; \ell)$ represents the output of the decoder for the point $p_i$ given the latent vector $\ell$.
        This works because sensor point cloud points lie on the surface of objects and have a real distance of $0$.
        As the \glspl{as} being reconstructed are not on the surface, this method can not be applied.
        Instead, we use PointNet++ as an encoder.
        As a decoder, we use a \gls{mlp} with a skip connection and batch normalization.
        An overview of the architecture is shown in \Cref{fig:architecture}.
        We use an aspect-ratio-preserving isotropic normalization on the input point clouds.
        Therefore, we compute a single scaling factor for all dimensions and a translation such that the point cloud fits inside $[-1,1]^3$.
        To decrease the sensitivity of the encoder to the point cloud density, we exclude up to $70\%$ of points obtained before encoding.
        This reduction also decreases training time, as encoding the point cloud is a bottleneck.
        As an additional augmentation during training, the data is randomly rotated by angles uniformly sampled (in degrees) from $\text{U}(-30, 30)\times\text{U}(-30, 30)\times\text{U}(-30, 30)$.

\newdimen\OccupancyNetworkX
\newdimen\OccupancyPointCloudY

\newcommand{\textnode}[3][2cm]{%
    \node (#3) [draw, rectangle, align=center, #2] {%
        \shortstack{%
            \parbox{#1}{\centering #3}%
        }%
    };%
}

\newcommand{\textnodedown}[2]{
    \node (#2) [draw, rectangle,align=center, #1] {
        \shortstack{
            \parbox{2cm}{\centering \rotatebox{90}{#2}}
        }
    };
}

\newcommand{\textnodedownmath}[3]{
    \node (#2) [draw, rectangle,align=center, #1] {
        \shortstack{
            \parbox{2cm}{\centering \rotatebox{90}{#3}}
        }
    };
}

\newcommand{\textnodemath}[3]{
    \node (#2) [draw, rectangle,align=center, #1] {
        \centering #3
    };
}

\newcommand{\textnodedownmathtinybound}[3]{
    \node (#2) [draw, rectangle, align=center, #1] {
            \centering \rotatebox{90}{\tiny #3}
    };
}

\newcommand{\randompoints}[1]{
        \begin{tikzpicture}
            \pgfmathsetseed{3245} %
            \foreach \i in {1,...,30} {
                \pgfmathsetmacro{\x}{rand*0.3} %
                \pgfmathsetmacro{\y}{rand*0.6} %
                \node[fill=black, circle, inner sep=0.5pt] at (\x, \y) {};
            }
            \pgfmathsetmacro{\x}{rand*0.3} %
            \pgfmathsetmacro{\y}{rand*0.6} %
            \node[fill=#1, circle, inner sep=1.5pt] at (\x, \y) {};
        \end{tikzpicture}
}

\newcommand{\mlp}[1]{
        \begin{tikzpicture}[node distance=0.1cm]
                \node (Input) [draw, rectangle,align=center]{\centering \rotatebox{90}{\tiny Input}};
                \node (A) [draw, rectangle,align=center, right=of Input]{\centering \rotatebox{90}{\tiny 1024 + 3}};
                \node (B) [draw, rectangle,align=center, right=of A]{\centering \rotatebox{90}{\tiny 512}};
                \node (C) [draw, rectangle,align=center, right=of B]{\centering \rotatebox{90}{\tiny 512}};
                \node (D) [draw, rectangle,align=center, right=of C]{\centering \rotatebox{90}{\tiny 512}};
                \node (E) [draw, rectangle,align=center, right=of D]{\centering \rotatebox{90}{\tiny 1024 + 3 + 512}};
                \node (F) [draw, rectangle,align=center, right=of E]{\centering \rotatebox{90}{\tiny 512}};
                \node (G) [draw, rectangle,align=center, right=of F]{\centering \rotatebox{90}{\tiny 512}};
                \node (H) [draw, rectangle,align=center, right=of G]{\centering \rotatebox{90}{\tiny 512}};
                \node (I) [draw, rectangle,align=center, right=of H]{\centering \rotatebox{90}{\tiny 67 + 1}};

                \draw[->] (Input) -- (A);
                \draw[->] (Input.south) to[bend right] (E.south);
                \draw[->] (A) -- (B);
                \draw[->] (B) -- (C);
                \draw[->] (C) -- (D);
                \draw[->] (D) -- (E);
                \draw[->] (E) -- (F);
                \draw[->] (F) -- (G);
                \draw[->] (G) -- (H);
                \draw[->] (H) -- (I);
        \end{tikzpicture}
}

\begin{figure}
    \centering
        \begin{tikzpicture}[node distance=0.2cm, auto]
          \begin{scope}[local bounding box=firstBox]
              
            \node (Sensor Point Cloud) [align=center] {
                \shortstack{
                    \includegraphics[width=0.8cm]{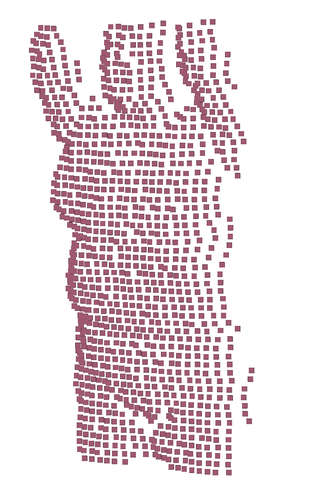}
                    \\
                    \parbox{2cm}{\centering Sensor Point Cloud}
                }
            };

            \node (Query Point) [right=of Sensor Point Cloud, align=center] {
                {\shortstack{\randompoints{black}\\Query Point\\$(x,y,z)$}}
            };

            \textnode[4cm]{below=of Sensor Point Cloud, xshift=1cm}{Isotropic Normalization}
            \textnode{below=of Isotropic Normalization, xshift=0.5cm}{Point Drop}
            \textnode{below=of Point Drop}{PointNet++}
            
            \textnodemath{below=of PointNet++}{Latent}{$(l_1,\cdots,l_{1024}) \times (x,y,z)$}
            
            \textnodemath{below=of Latent}{MLP}{\shortstack{\mlp\\\\MLP}}
            \textnodemath{below=of MLP}{Output}{\shortstack{\randompoints{magenta}\\\textcolor{magenta}{Occupancy} + Signed Distance}}
            \textnode[5cm]{below=of Output, yshift=-0.05cm}{Isotropic Denormalization}

            \draw[->, thick, >=stealth] (Sensor Point Cloud.south) -| (Isotropic Normalization);
            \draw[->, thick, >=stealth] (Isotropic Normalization.south) -- ($(Point Drop.north) - (0.5cm, 0.0cm)$);
            \draw[->, thick, >=stealth] (Point Drop) -- (PointNet++);
            \draw[->, thick, >=stealth] (Query Point.south) -| node[pos=0.8, sloped, above] {\tiny{Append}} (Latent.north east);
            \draw[->, thick, >=stealth] (PointNet++) -- (Latent);
            \draw[->, thick, >=stealth] (Latent) -- (MLP);
            \draw[->, thick, >=stealth] (MLP) -- (Output);
            \draw[->, thick, >=stealth] (Output) -- (Isotropic Denormalization);

            \draw[->, thick, >=stealth] (Isotropic Normalization.west) -- +(-0.9cm,0) |- node[pos=0.25, sloped, above]{$(\text{scale}, \text{translation}) \in\mathbb{R}\times\mathbb{R}^3 $} (Isotropic Denormalization.west);

            \node (Container) [draw, black, fit=(Point Drop) (PointNet++) (Latent) (MLP) (Output), inner sep=0.15cm, dashed] {};
            \node (Occupancy Network) at (Container.north west) [below, inner sep=2mm, anchor=north west]{ %
                \shortstack{
                    \includegraphics[height=0.8cm]{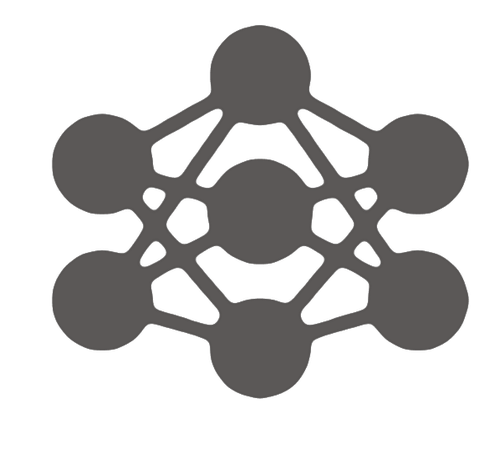}\\
                    \parbox{1cm}{\centering \tiny Occupancy Network}
                }
            };
        
          \end{scope}
        \end{tikzpicture}
    \caption{
        The \textit{Sensor Point Cloud} is normalized and passed to a \textit{Point Drop} node, that randomly discards up to $70\%$ of all points.
        The remaining points are the input to \textit{PointNet++} which distills a latent vector $(l_1,\cdots,l_{1024})$.
        The latent vector is appended with a query point $(x,y,z)$.
        The combined vector is used as input to the \textit{MLP}.
        Additionally, a skip connection to the layer $5$ is used.
        Between each layer, ReLU is used as an activation function.
        During training, batch normalization is used for the hidden layers of the \textit{MLP}.
        The output is the predicted occupancy value (one-hot) and the distance to the signed nearest surface.
        The output is denormalized to obtain the output in the original camera coordinate system.
    }
    \label{fig:architecture}
\end{figure}

        Our loss function is a combination of a cross-entropy loss and a distance loss, defined as $\mathcal{L} = \text{CE}(o,\hat{o}) + \lambda\,\|d - \hat{d}\|_2^2.$
        We set $\lambda = 100$ to balance the two terms, ensuring that they are of similar scale.
        For optimization, we use the Adam optimizer with a learning rate of $0.0005$.

    \subsection{Revised SortSample}

        Henrich et al.~\cite{henrich2024registered} introduced SortSample as an algorithm to produce training data for occupancy or signed distance field learning.
        SortSample, applied to this application, works as follows:
            \begin{enumerate}
                \item Sample points uniformly within each \gls{as}'s 50\% enlarged bounding box.
                \item Assign each sampled point to either set $S_i$ if inside or $S_o$ if outside of the \gls{as}.
                \item Remove points in $S_o$ that are also inside another \gls{as}.
                \item Continue drawing samples until $min(|S_i|, |S_o|) = N$, where $N$ is a hyper parameter.
                \item Sort $S_i$ and $S_o$, using the distance to the nearest surface as the sorting key.
                \item Only keep the closest $N$ points from $S_i$ and $S_o$.
            \end{enumerate}

        This approach results in training data that can effectively train occupancy or SDF networks to represent objects consisting of similarly sized segments.
        However, SortSample has limitations.
            It does not allow the learning of tightly packed segments, as each segment requires outside samples.
            As the inside of the human body is tightly packed, no outside labels exist for internal \gls{as}.
            Therefore, $S_o$ will never be populated and SortSample will not terminate.
            Although SortSample may not be biased for single \glspl{as}, for multiple touching \glspl{as}, there is a local bias towards the smaller and more densely sampled \gls{as}.
            This results in a ballooning effect of the smaller structure into the larger structure.

        We propose an effective fix for the SortSample algorithm.
            SortSample discards points that are inside of other \glspl{as}.
            We redefine $S_o$ to not only contain points outside of all \gls{as}.
            Instead, $S_o$ contains all points outside of the currently sampled \gls{as}.
            This includes points inside of other \glspl{as}.
            After sorting the sets and keeping only the closest $N$ points, all points in $S_o$ are assigned the label of the \gls{as} that they are inside of.
            Therefore, the inside points of embedded structures are enclosed by equally densely sampled points inside of other \gls{as}.
            This allows the representation learning of tightly packed structures.
            Further, it eliminates the local density bias stated in the original work~\cite{henrich2024registered}.

        \subsection{Training Data}
            \label{sec:training_data}
            To generate the training dataset, we use the Atlas Dataset~\cite{jaus2023towards} containing anatomical masks for $533$ CT scans with a total of 142 distinguishable \glspl{as}. 
            For generating the training data, only the masks are needed.
            
            Certain \glspl{as} are consistently missing or degenerate across multiple masks. We exclude these structures from our training and evaluation data.
            This also excludes sex-specific organs, such as the ovaries and the prostate.
            Further, we grouped the remaining \glspl{as} to create more coarse objects.
            For example, Atlas Dataset separates all vertebrae, which we group into cervical, lumbar, thoracic, and sacrum.
            After filtering and grouping, $67$ \glspl{as} remain, see \Cref{fig:combined}.
            Additionally, $36$ masks were removed from the original dataset of $533$ due to missing structures.
            Of the remaining $497$ masks, $50$ are reserved for evaluation.
            Therefore, $447$ different masks are used for training.

\newdimen\OccupancyNetworkX
\newdimen\OccupancyPointCloudY

\newcommand{\graphicnode}[3]{

    \node (#2) [align=center, #1] {
        \shortstack{
            \includegraphics[height=2cm]{#3} \\
            \parbox{2cm}{\centering #2}
        }
    };
}

\newcommand{\textnodeplf}[2]{

    \node (#2) [draw, rectangle, align=center, #1] {
        \shortstack{
            \parbox{2cm}{\centering #2}
        }
    };
}

\begin{figure}
    \centering
    \resizebox{0.9\columnwidth}{!}{
        \begin{tikzpicture}[node distance=0.25cm, auto]
          \begin{scope}[local bounding box=firstBox]
              
        \node (Atlas Dataset) [align=center] {\includegraphics[width=2cm]{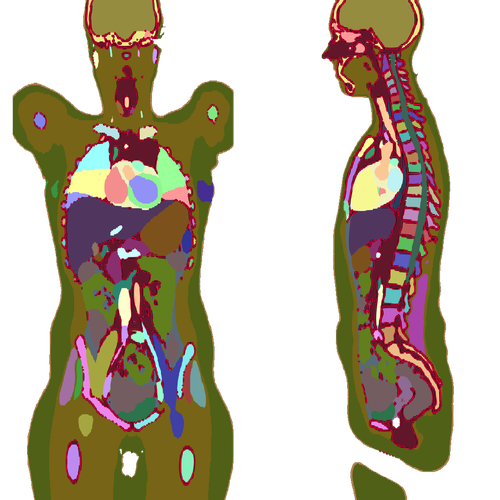}\\Atlas Dataset};
        \graphicnode{right=of Atlas Dataset}{Extract Meshes}{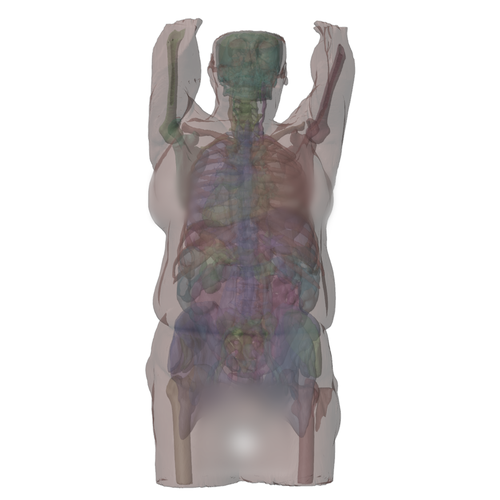}
        \graphicnode{below=of Extract Meshes}{Augmentation}{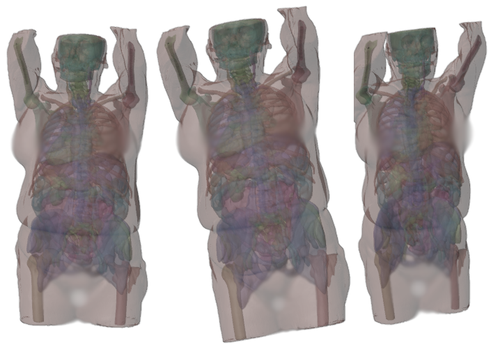}
        \graphicnode{below=of Augmentation}{Query Point Cloud}{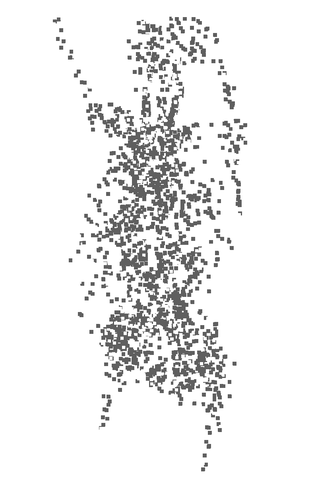}
        \graphicnode{left=of Query Point Cloud}{Sensor Point Cloud}{graphics/pipeline_figure/surface_point_cloud.png}
        \graphicnode{right=of Query Point Cloud}{Occupancy Samples}{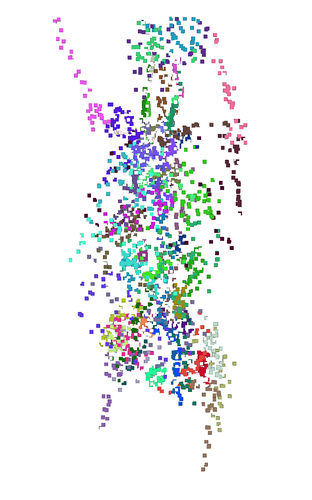}
        
        \graphicnode{below=1.2cm of Query Point Cloud}{Occupancy Network}{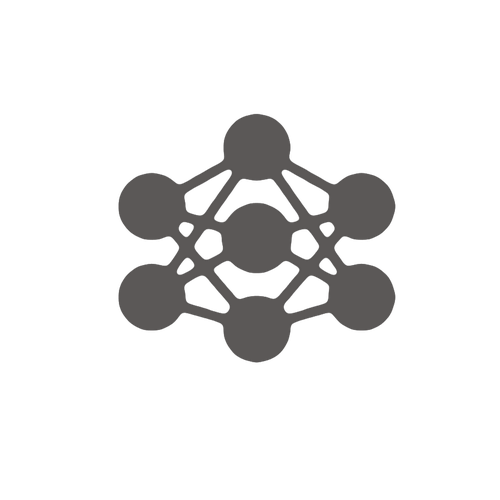}
        
        \graphicnode{right=of Occupancy Network}{Estimated Labels}{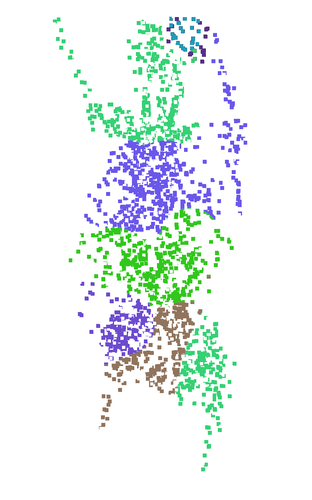}
        \textnodeplf{above=of Estimated Labels}{CE + L2 Loss}
        
        \draw[->] (Atlas Dataset) -- (Extract Meshes);
        \draw[->] (Extract Meshes) -- (Augmentation);
        \draw[->] (Augmentation.east) to[bend left] (Occupancy Samples.north);
        \draw[->] (Augmentation.west) to[bend right] (Sensor Point Cloud.north);
        \draw[->] (Occupancy Samples) -- (Query Point Cloud);
        \draw[->] (Query Point Cloud) -- (Occupancy Network);
        
        \draw[->] (Sensor Point Cloud.south) to[bend right] (Occupancy Network.west);
        \draw[->] (CE + L2 Loss.west) to[bend right] node[pos=0.5, sloped, below] {\footnotesize	 Update} (Occupancy Network);
        \draw[->] (Occupancy Network) -- (Estimated Labels);
        \draw[->] (Estimated Labels) -- (CE + L2 Loss);
        \draw[->] (Occupancy Samples) -- (CE + L2 Loss);
        
          \end{scope}
        \end{tikzpicture}
      }
    \caption{The data and training pipeline. A random mask is selected from the \textit{Atlas Dataset}. Anatomical structures are grouped and meshes are extracted. To produce training data, all meshes obtained from a mask are augmented through deformations and camera movements. The improved SortSample is used to obtain an \textit{Occupancy Samples}. Simultaneously, a \textit{Sensor Point Cloud} from the camera perspective is created. The \textit{Query Point Cloud} is obtained by removing class information from the \textit{Occupancy Samples}. The \textit{Occupancy Network}, conditioned on the \textit{Sensor Point Cloud}, estimates the labels for all point in the \textit{Query Point Cloud}. The loss is computed with respect to the \textit{Occupancy Samples} and used to update the \textit{Occupancy Network}.}
    \label{fig:pipeline}
\end{figure}

            To prepare the Atlas Dataset for our training, we apply the same processing steps to each mask, a general overview is shown in \Cref{fig:pipeline}.
            For each \gls{as}, we obtain a surface mesh representation using marching cubes~\cite{lorensen1998marching}.
            As the original representation is volumetric, where each point is assigned exactly one class, there can be no intersections (or self-intersections).
            To ensure that all meshes are watertight (i.e., without holes), we add a 1 voxel thick boundary around each mask volume. 
            Atlas Dataset contains some floating regions, wrongly classified.
            Consequently, for \glspl{as} that should form a single connected entity, we select the largest connected mesh.
            This process effectively eliminates visual floating artifacts.
            Some \glspl{as}, such as the ribs and rib cartilage, also contain many floating artifacts.
            As the ribs are not a large connected entity, which allow largest connected mesh filtering, these are removed manually.
            Additionally, we extract the skin surface by grouping all \glspl{as}.
            Notice, performing marching cubes directly on the skin label would result in a mesh with a "thickness", as the inside and the outside of the skin would be converted to surfaces.
            
            To generate a training example, which consists of a sensor point cloud (derived from a depth image) and occupancy sample pair, the following procedure is followed:
            A random mask is selected.
            All meshes belonging to the mask are loaded into a virtual scene and a virtual camera is placed at a random distance (meters) ${d \sim \text{U}(1.4, 2.6)}$.
            Additionally, the camera is moved randomly on the orthogonal plane of the viewing direction by the horizontal and vertical offsets ${h \sim \text{U}(-0.7, 0.7)}$ and ${v \sim \text{U}(-0.1, 0.3)}$.
            The camera is then pointed towards the center of the mask.
            Adjusting the camera position is needed to reflect that the exact distance or relative position of a patient to the camera is often unknown.
            
            A $64\times64$ depth image is captured of the skin and the revised SortSample ($N=32$) is used to obtain $32$ inside and $32$ outside occupancy samples for each \gls{as}.
            This excludes the skin itself, which is not reconstructed.
            The depth image is converted to a point cloud, as we use PointNet++~\cite{qi2017pointnet++} as an encoder.
            The occupancy samples are then transformed into the camera space.
            
            A total of $31\,984$ point cloud and occupancy sample pairs are used for training.
            Early tests on the $447$ masks resulted in low generalizability.
            As a result, real-world data resulted in no organs being reconstructed.

\begin{figure}[t]
    \centering
    \resizebox{0.75\columnwidth}{!}{
        \begin{tikzpicture}
            \node (img1) at (0,0) {
                \begin{adjustbox}{valign=t}
                \begin{tikzpicture}
                  \node[anchor=south west,inner sep=0] (image) at (0,0) {\includegraphics[width=0.2\textwidth]{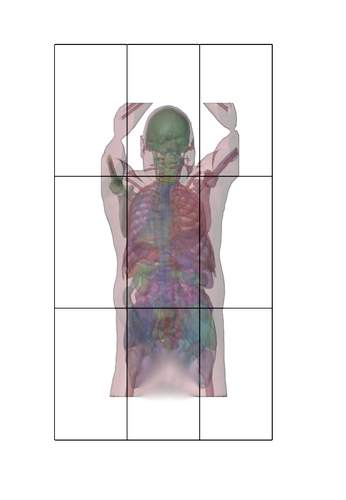}};
                  \begin{scope}[x={(image.south east)},y={(image.north west)}]
                    \draw[fill=red] (0.16,0.91) circle (2pt) node[anchor=north west] {$1$};
                    \draw[fill=red] (0.16+0.205,0.91) circle (2pt) node[anchor=north west] {$2$};
                    \draw[fill=red] (0.16+0.205*2,0.91) circle (2pt) node[anchor=north west] {$3$};
                    \draw[fill=red] (0.16+0.205*3,0.91) circle (2pt) node[anchor=north west] {$4$};
                  \end{scope}
    
                  \begin{scope}[shift={(1.7cm, 0.4cm)},scale=0.15]
                    \tdplotsetmaincoords{60}{90}
                    \begin{scope}[tdplot_main_coords]
                      \coordinate (O) at (0,0,0);
                      \coordinate (X) at (0,-2.0,0);
                      \coordinate (Y) at (1.5,0,0);
                      \coordinate (Z) at (0,0,2.0);
    
                      \draw[->,red,thick] (O) -- (X) node[anchor=east]{$X$};
                      \draw[->,green,thick] (O) -- (Y) node[anchor=west]{$Y$};
                      \draw[->,blue,thick] (O) -- (Z) node[anchor=south]{$Z$};
                    \end{scope}
                  \end{scope}
                \end{tikzpicture}
                \end{adjustbox}
            };
            \node[below=2pt of img1] {(a)};
    
            \node (img2) [right=0.4cm of img1] {\includegraphics[width=0.2\textwidth]{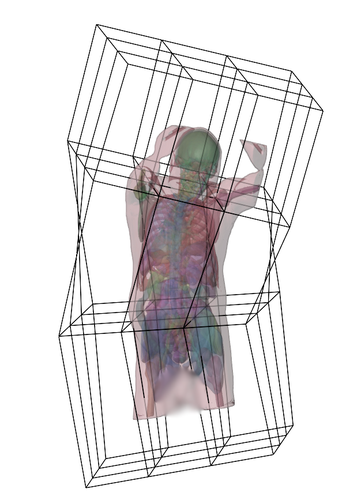}};
            \node[below=2pt of img2] {(b)};
    
            \node (img3) [below=0.4cm of img1] {\includegraphics[width=0.2\textwidth]{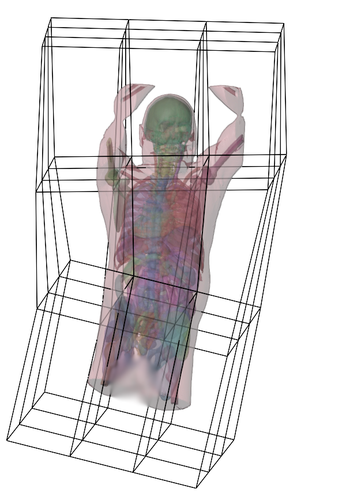}};
            \node[below=2pt of img3] {(c)};
    
            \node (img4) [below=0.4cm of img2] {\includegraphics[width=0.2\textwidth]{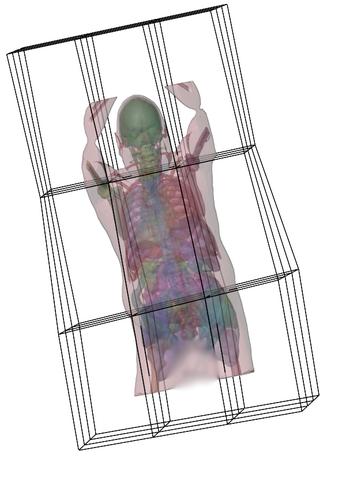}};
            \node[below=2pt of img4] {(d)};
    
        \end{tikzpicture}
    }
    \caption{Examples of the mask data augmentation using a $4\times4\times4$ lattice (\textcolor{red}{red dots} in (a)). The original mask (a) is augmented using the lattice to obtain augmentations (b), (c), and (d).}
    \label{fig:deformation_examples}
\end{figure}

            To augment the data and include more diverse poses of the human body, we used a $4\times4\times4$ lattice based deformation.
            Each level of the lattice is scaled and rotated to deform the meshes used to produce the training data, see \cref{fig:deformation_examples}.
            The lattice vertices are positioned such that the middle two levels are at approximately the hip and shoulders.
            The origin of the rotations and scaling is the geometric center of the lattice level.
            The rotations are sampled (in degrees) from $\text{U}(-15, 15)\times\text{U}(-10, 10)\times\text{U}(-15, 15)$ for the top and bottom layer.
            For the middle levels, they are sampled from $\text{U}(-5, 5)\times\text{U}(-5, 5)\times\text{U}(-2.5, 2.5)$.
            The spatial axes are shown in \Cref{fig:deformation_examples}.
            The bottom and top most levels are only moved by a small amount, as these affect more rigid structures that should not deform as much.
            The scaling factor in each level is sampled from $\text{U}(0.85, 1.15)$.

        \subsection{Evaluation Data}
            For the evaluation dataset, we use the first $50$ valid masks without any deformation or rotational augmentation.
            We compute an \gls{aabb} for each \gls{as}.
            This bounding box represents the \gls{roi} that we aim to estimate.
            
            Additionally, we collect real-world point clouds of 12 individuals ($7$ female, $5$ male) for a qualitative reconstructions with real-world data.
            Ages ranged between $26$ and $40$, height between $154\text{cm}$ and $194\text{cm}$, weight between $55\text{kg}$ and $97\text{kg}$.
            All individuals were fully clothed, but loose clothing was pulled tight.
            This is in contrast to the training data, where all masks are obtained from individuals without clothing.
            We use a PMD flexx2 3D camera (pmdtechnologies AG, Germany).
            To remove the background, all points further than $2.5$ meters are discarded.
            Finally, the point cloud is cropped, excluding the lower legs and the forearms to match the dataset, see \Cref{fig:overview}.
            The PMD flexx2 produces denser point clouds than needed, thus, point clouds are sub-sampled randomly until $1000$ points remain.
            
        \subsection{Template Matching Baseline}
            We use template matching as a baseline.
            For this, we consider all $447$ masks from the training dataset as templates.
            For each template mask, a frontal sensor point cloud is captured using the same configuration as for the evaluation data.
            Therefore, no deformations or rotations are applied.
            This is done to ensure the template matching performs optimally.
            To infer the position of anatomical structures, the patient sensor point cloud is registered to each of the template point clouds.
            We use \gls{icp} for the registration.

            The registered mask with the minimal chamfer distance is used as the template to locate each anatomical structure and provides the \glspl{aabb}.
            Note that while \gls{icp} only handles rigid transformations, using the set of different template masks addresses body-shape variability.
    
        \subsection{Evaluation Method}

\begin{figure}
    \centering
    \vspace{2mm}

        \centering
        \begin{tikzpicture}[scale=0.4]

        \begin{scope}[shift={(0,0)}]
            \draw[dashed] (-0.25,-0.25) rectangle ++(2.5,2.5);
            \draw[fill=blue!30] (0,0) rectangle ++(2,2);
            \draw[fill=red!30] (0.25, 0.25) rectangle ++(2,2);
            \draw[->, thick] (2,0) -- (2.25-0.05,-0.25+0.05);

            \node[anchor=south] at (1.125,2.2) {$2$~cm};
            \node[anchor=south, rotate=-90] at (2.2,1.125) {$2$~cm};
            
            \node[anchor=south east] at (3.25,-3) {$\text{IoU} = 0.62$};
            \node[anchor=south east] at (3.25,-4) {$\text{ESF} = 1.25$};
            \node[anchor=south east] at (3.25,-2) {$\text{CD} = 0.35$};
            \node[anchor=south west, font=\scriptsize] at (2.66,-2) {$\tiny\text{cm}$};

            \fill[blue!70] (1,1) circle (4pt); %
            \fill[red!70] (1.25,1.25) circle (4pt); %
            \draw[dashed] (1,1) -- (1.25,1.25);
        \end{scope}
        
        \begin{scope}[shift={(5,0)}]
            \draw[dashed] (-0.5,-0.5) rectangle ++(3,3);
            \draw[fill=blue!30] (0,0) rectangle ++(2,2);
            \draw[fill=red!30] (0.5,0.5) rectangle ++(2,2);
            \draw[->, thick] (2,0) -- (2.5-0.25,-0.5+0.25);

            \node[anchor=south east] at (3.25,-3) {$\text{IoU} = 0.39$};
            \node[anchor=south east] at (3.25,-4) {$\text{ESF} = 1.50$};
            \node[anchor=south east] at (3.25,-2) {$\text{CD} = 0.71$};
            \node[anchor=south west, font=\scriptsize] at (2.66,-2) {$\tiny\text{cm}$};

            \fill[blue!70] (1,1) circle (4pt); %
            \fill[red!70] (1.5,1.5) circle (4pt); %
            \draw[dashed] (1,1) -- (1.5,1.5);

        \end{scope}
        
        \begin{scope}[shift={(10,0)}]
            \draw[dashed] (-0.75,-0.75) rectangle ++(3.5,3.5);
            \draw[fill=blue!30] (0,0) rectangle ++(2,2);
            \draw[fill=red!30] (0.75,0.75) rectangle ++(2,2);
            \draw[->, thick] (2,0) -- (2.75-0.25,-0.75+0.25);

            \node[anchor=south east] at (3.25,-3) {$\text{IoU} = 0.24$};
            \node[anchor=south east] at (3.25,-4) {$\text{ESF} = 1.75$};
            \node[anchor=south east] at (3.25,-2) {$\text{CD} = 1.06$}; 
            \node[anchor=south west, font=\scriptsize] at (2.66,-2) {$\tiny\text{cm}$};

            \fill[blue!70] (1,1) circle (4pt); %
            \fill[red!70] (1.75,1.75) circle (4pt); %
            \draw[dashed] (1,1) -- (1.75,1.75);
            
        \end{scope}
        
        \begin{scope}[shift={(15,0)}]
            \draw[dashed] (-1,-1) rectangle ++(4,4);
            \draw[fill=blue!30] (0,0) rectangle ++(2,2);
            
            \draw[fill=red!30] (1.0,1.0) rectangle ++(2,2);
            \draw[->, thick] (2,0) -- (3-0.25,-1+0.25);

            \node[anchor=south east] at (3.25,-3.0) {$\text{IoU} = 0.14$};
            \node[anchor=south east] at (3.25,-4) {$\text{ESF} = 2.00$};
            \node[anchor=south east] at (3.25,-2) {$\text{CD} = 1.41$};
            \node[anchor=south west, font=\scriptsize] at (2.66,-2) {$\tiny\text{cm}$};

            \fill[blue!70] (1,1) circle (4pt); %
            \fill[red!70] (2.0,2.0) circle (4pt); %
            \draw[dashed] (1,1) -- (2.0,2.0);

        \end{scope}

        \end{tikzpicture}

    \caption{Visual examples for 2D CE, IoU and ESF. The \textcolor{red}{red} box is the target. The \textcolor{blue}{blue} box is the estimate. Both are $2\times2$cm. The length of the line connecting the box centers is the CD. IoU decreases as the overlap between the boxes decreases. ESF increases as the \textcolor{blue}{estimated box} needs to be scaled up to encompass the \textcolor{red}{target box}.}
    \label{fig:iou_and_esf}
\end{figure}

            For all occupancy network evaluations, we reconstruct the 3D atlas at a volumetric resolution of $500\times500\times500$ to ensure small structures are not missed.
            We use a basic hierarchical sampling approach to improve inference time.
            For this, we first query the occupancy network with $40\,000$ random query points.
            Then, only the bounding box enlarged by $15\%$ of the points that lie inside are densely sampled.
            The predicted atlas is then used to compute the \gls{aabb} of each \gls{as}.
            The reference \glspl{aabb} are obtained directly from the raw masks.
            
            We employ three metrics to assess the accuracy of our estimates for the locations of \glspl{as}.
            The metrics are visualized in \Cref{fig:iou_and_esf}.
            The \gls{aabb} \gls{cd} measures how well the position of the \gls{as} is estimated.
            For this, the \gls{aabb} of the generated \gls{as} is compared against the reference bounding box:
            $\text{CD}(A, B) = \big|\big| A_{\text{center}} - B_{\text{center}}\big|\big|$.
            
            The \gls{aabb} \gls{iou} measures how well the size of the \gls{as} is estimated:
            $\text{IoU}(A, B) = \text{Volume}(A \cap B) \div \text{Volume}(A \cup B)$.
            
            The \gls{aabb} \gls{esf} measures how much the estimated \gls{aabb} needs to be scaled such that the reference \gls{aabb} is fully contained in it.
            If the reference is already contained, the \gls{esf} is $1$.
            In practice, \gls{esf} measures how much an estimated \gls{aabb} should be scaled to ensure that the \gls{as} of interest is fully captured by an imaging device during CT or MRI.
    \section{Results and Discussion}

\makeatletter
\newcommand{\filename}[1]{out_000\two@digits{#1}.png}
\makeatother

\newlength{\sfactor}
\setlength{\sfactor}{0.12\textwidth}

\begin{figure}
\centering
  \includegraphics[width=0.9127\sfactor, trim={3cm 11cm 23cm 1cm},clip]{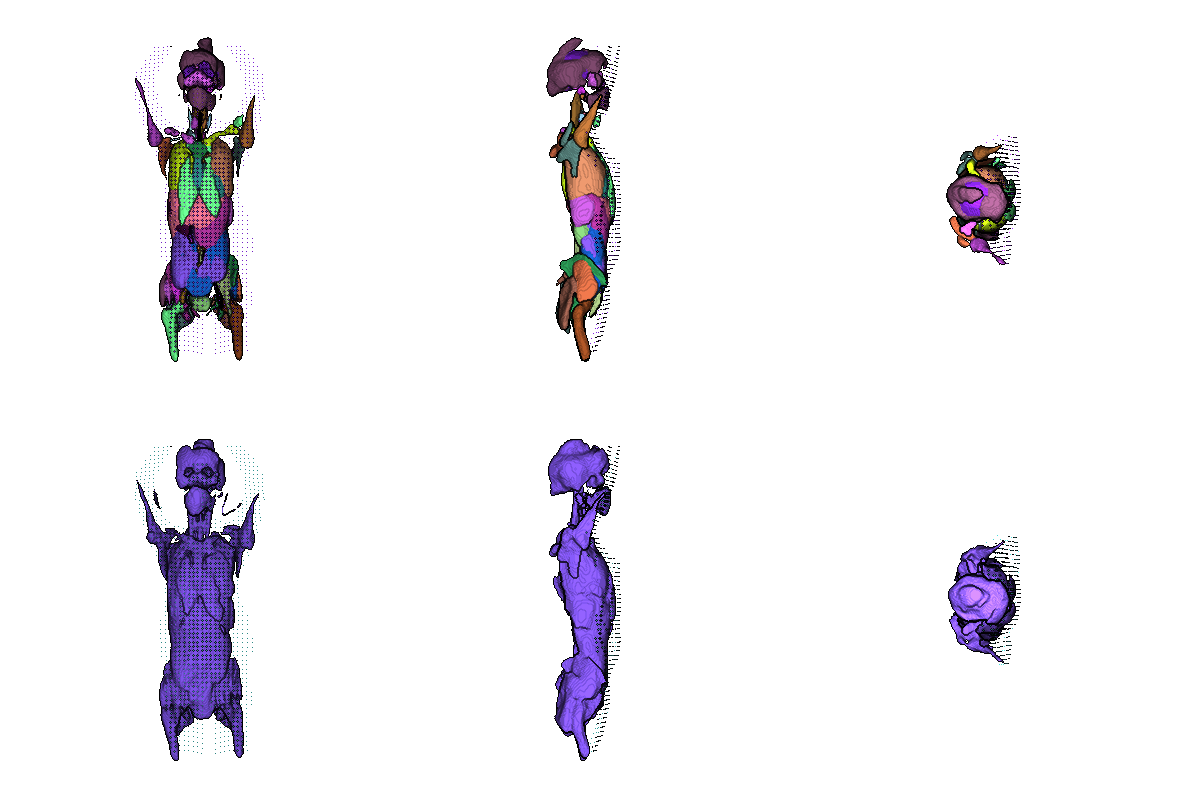}
  \includegraphics[width=0.9996\sfactor, trim={3cm 11cm 23cm 1cm},clip]{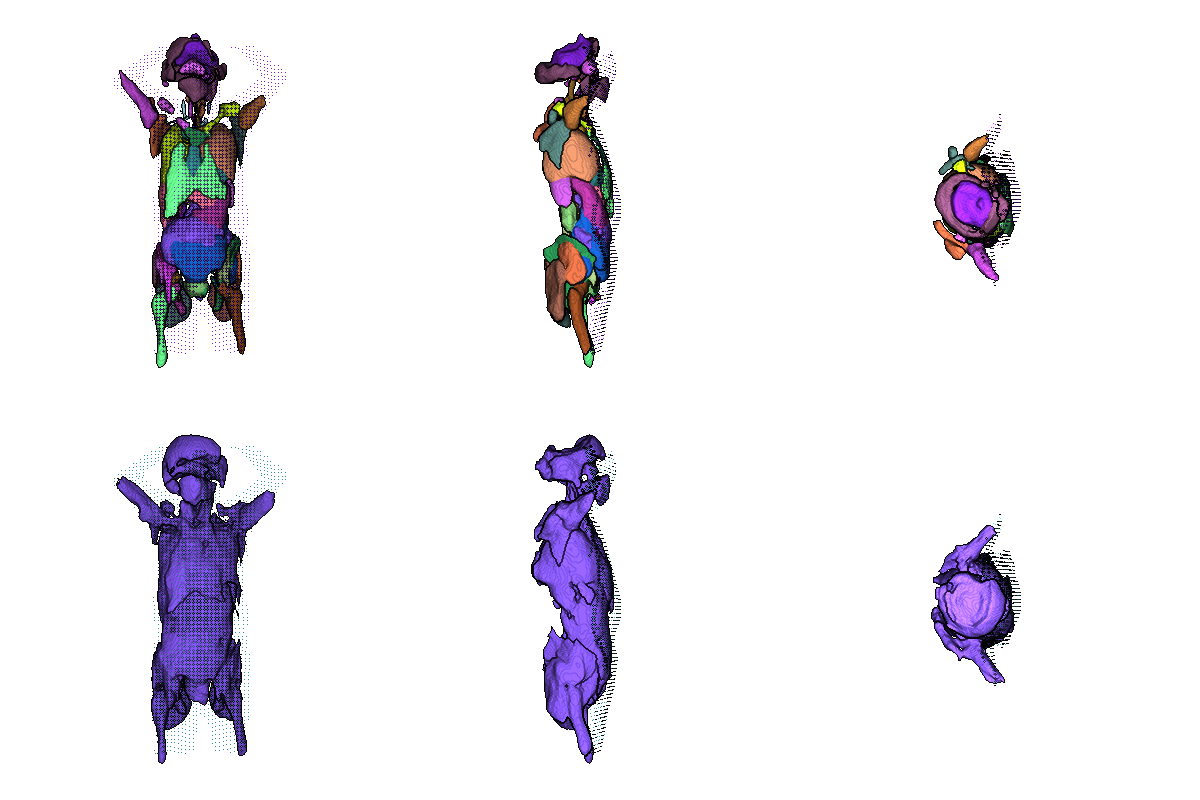}
  \includegraphics[width=0.8936\sfactor, trim={3cm 11cm 23cm 1cm},clip]{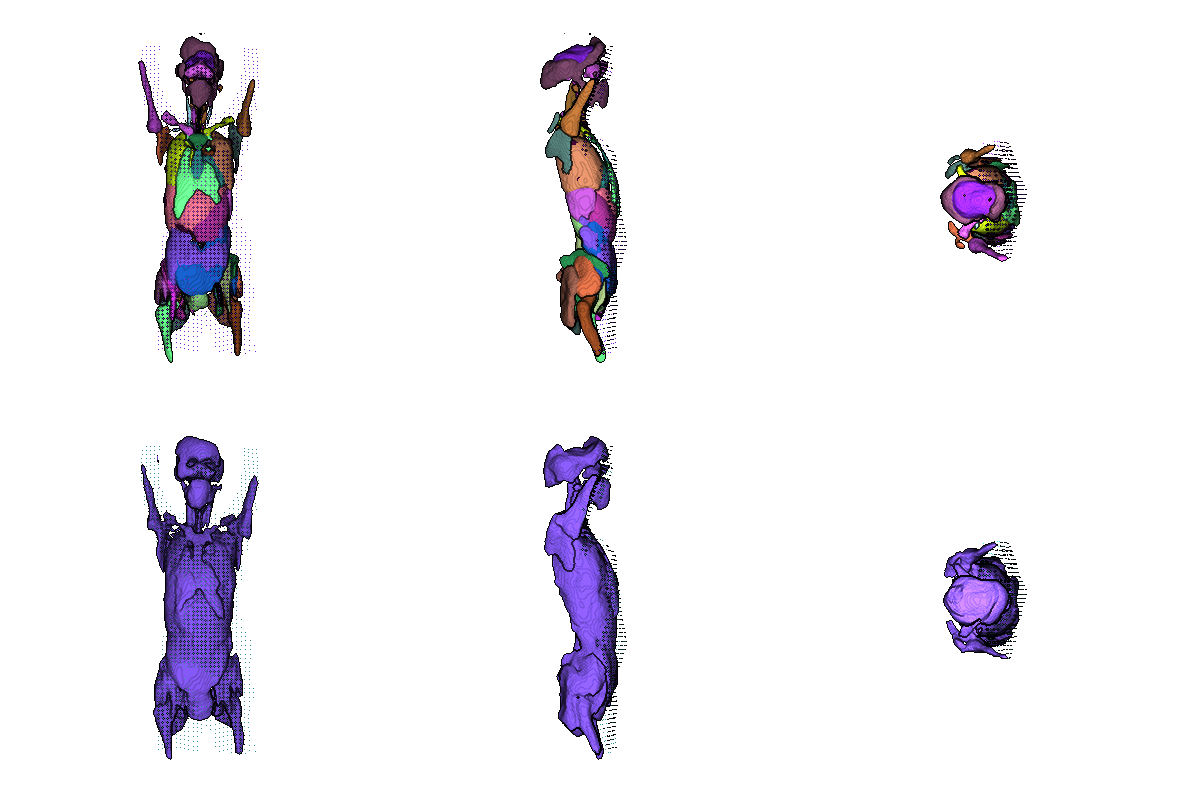}
  \includegraphics[width=0.8791\sfactor, trim={3cm 11cm 23cm 1cm},clip]{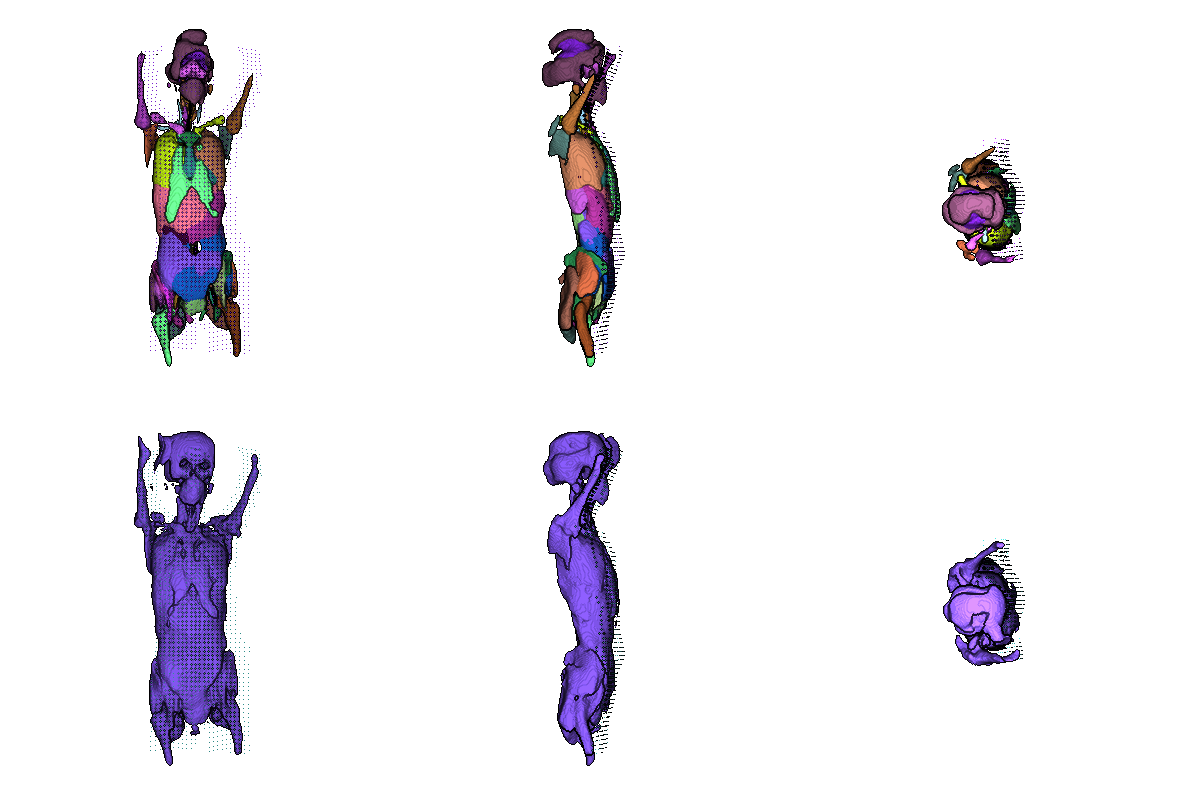}
  \includegraphics[width=0.9513\sfactor, trim={3cm 11cm 23cm 1cm},clip]{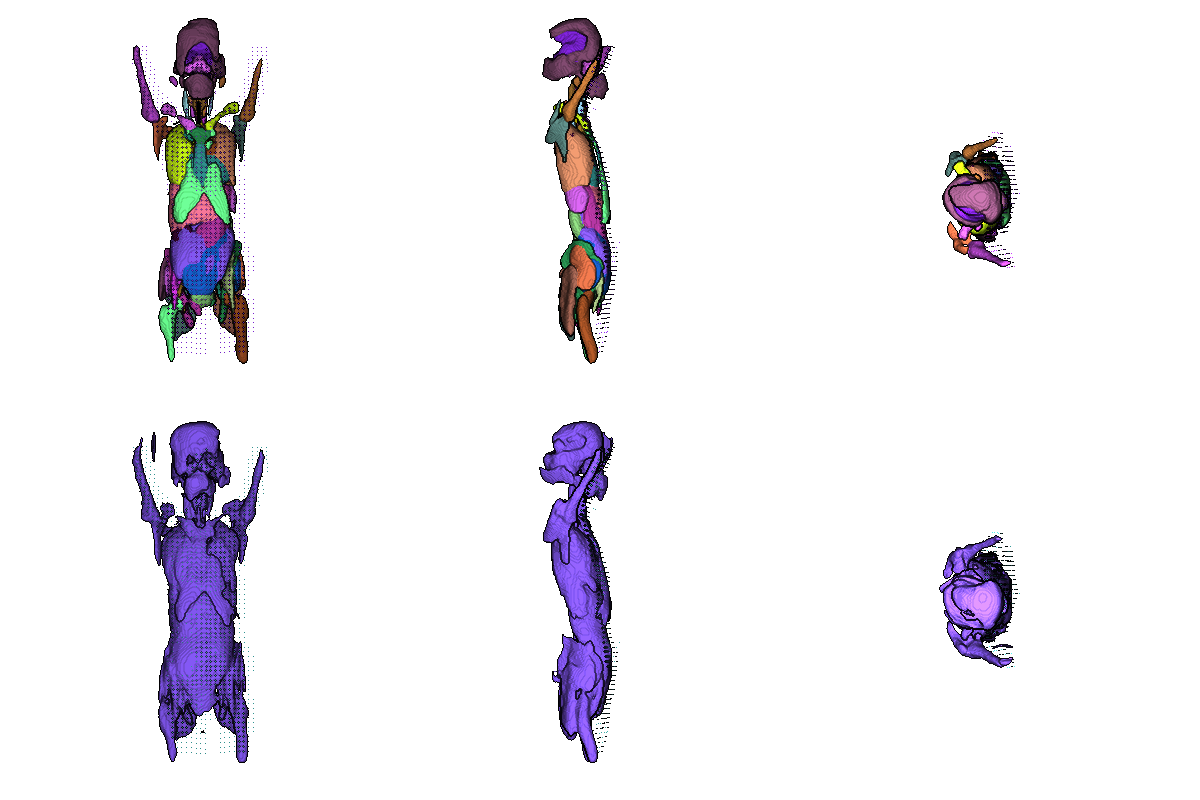}
  \includegraphics[width=0.8548\sfactor, trim={3cm 11cm 23cm 1cm},clip]{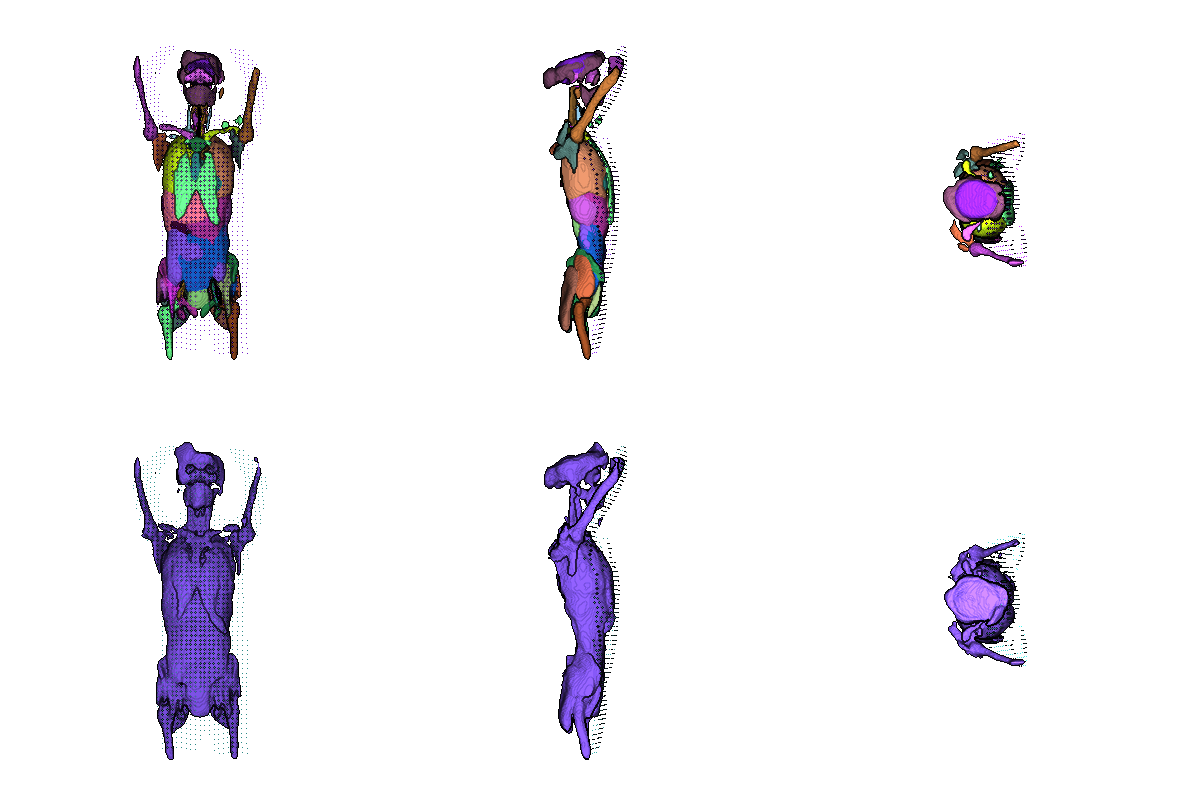}
  \includegraphics[width=0.9454\sfactor, trim={3cm 11cm 23cm 1cm},clip]{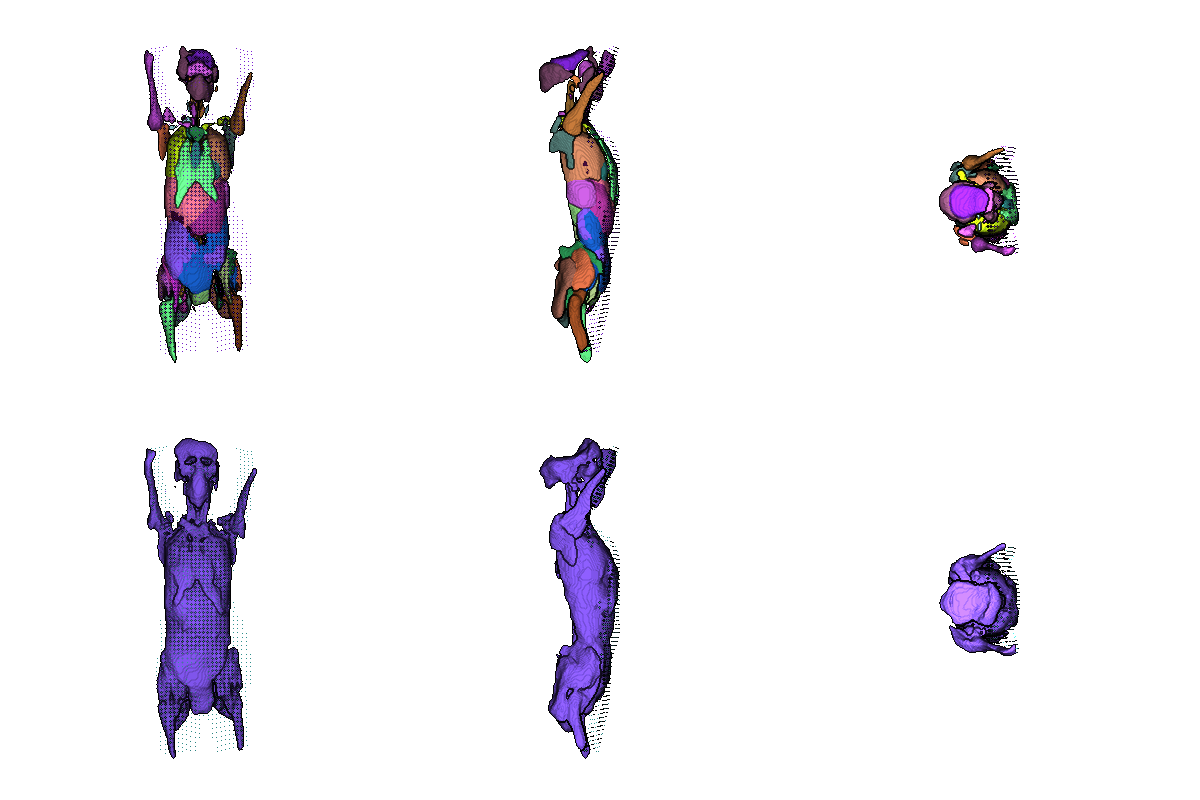}
  \includegraphics[width=1.0000\sfactor, trim={3cm 11cm 23cm 1cm},clip]{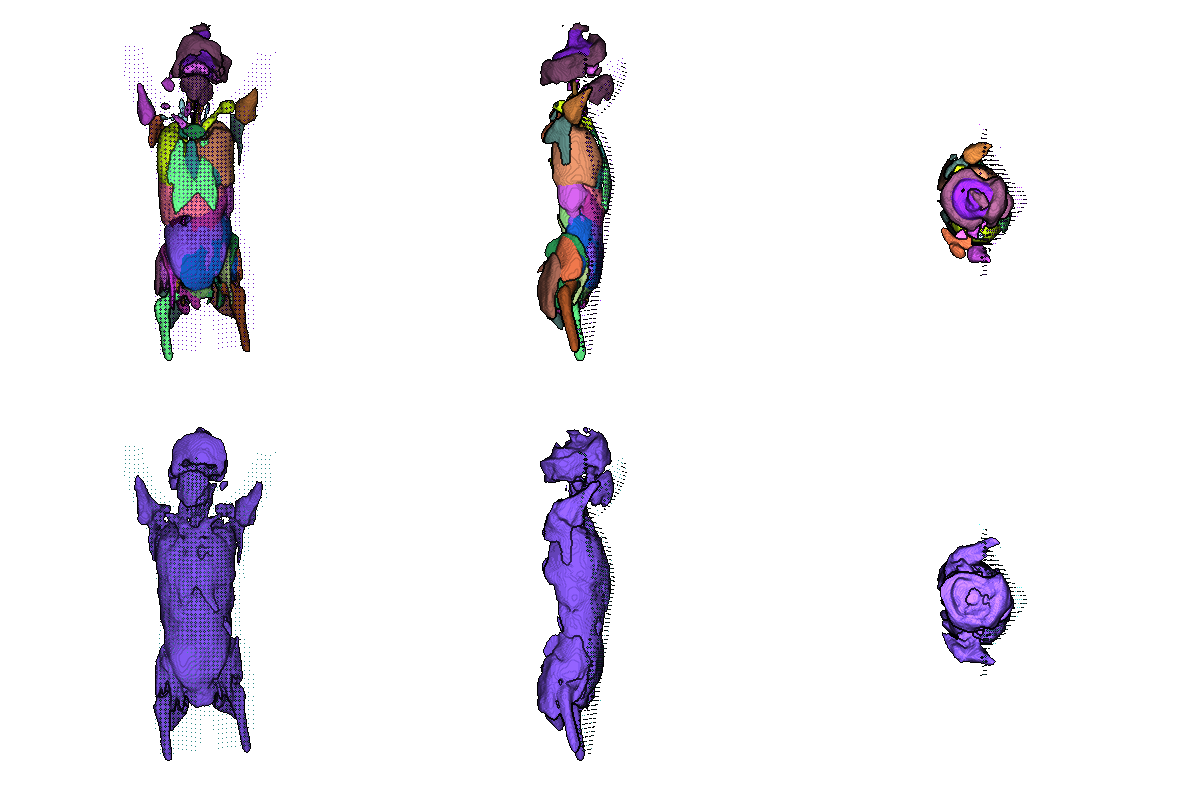}
  \includegraphics[width=0.8221\sfactor, trim={3cm 11cm 23cm 1cm},clip]{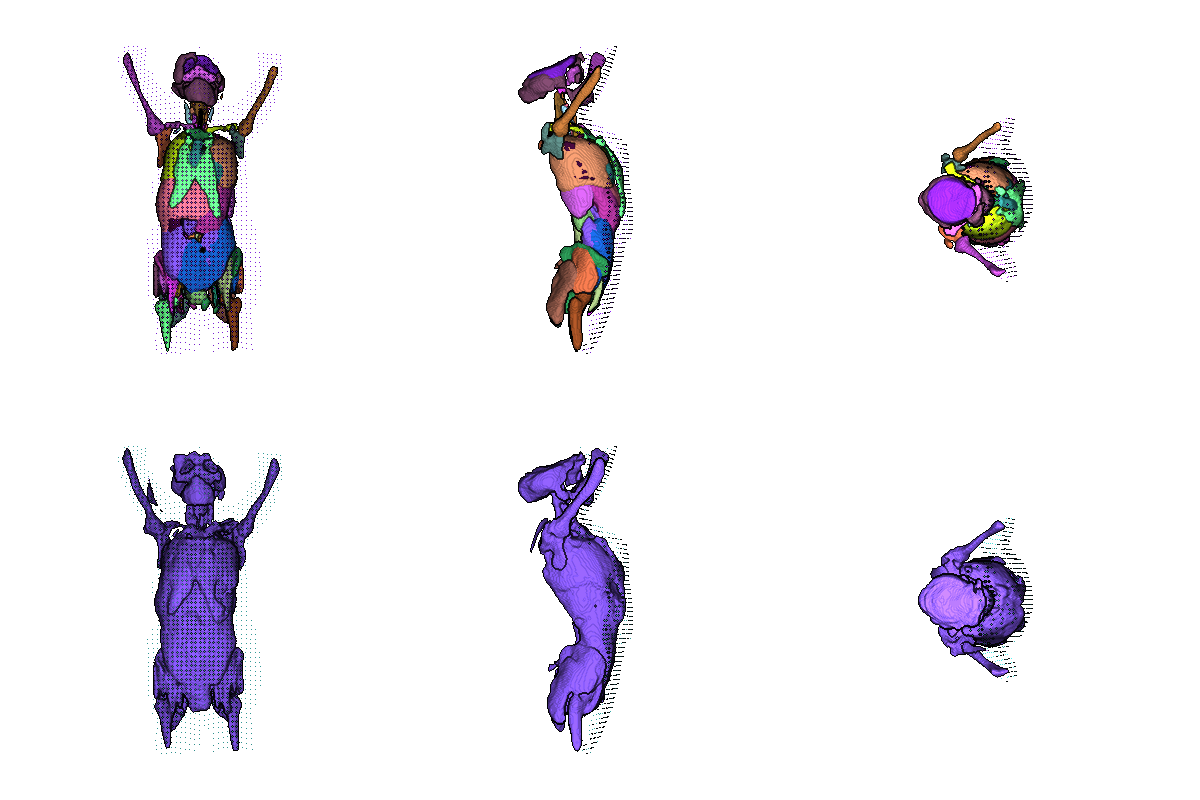}
  \includegraphics[width=0.8401\sfactor, trim={3cm 11cm 23cm 1cm},clip]{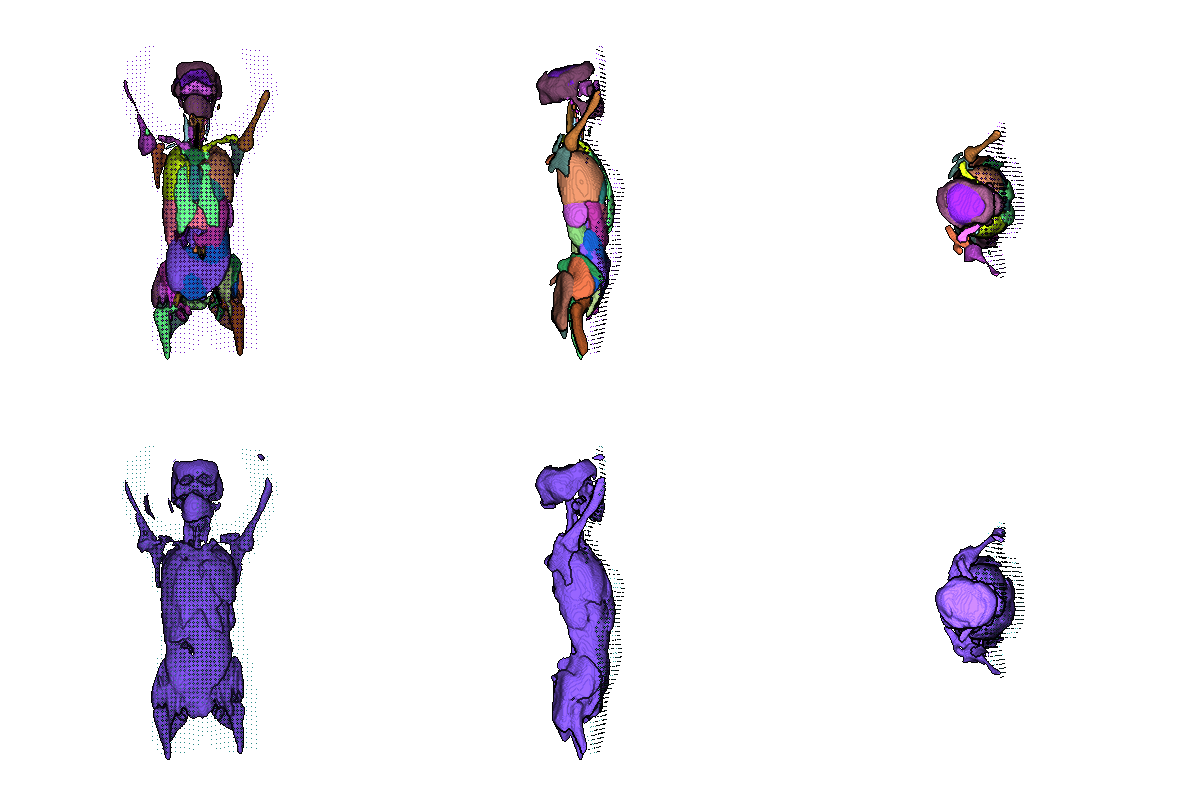}
  \includegraphics[width=0.9129\sfactor, trim={3cm 11cm 23cm 1cm},clip]{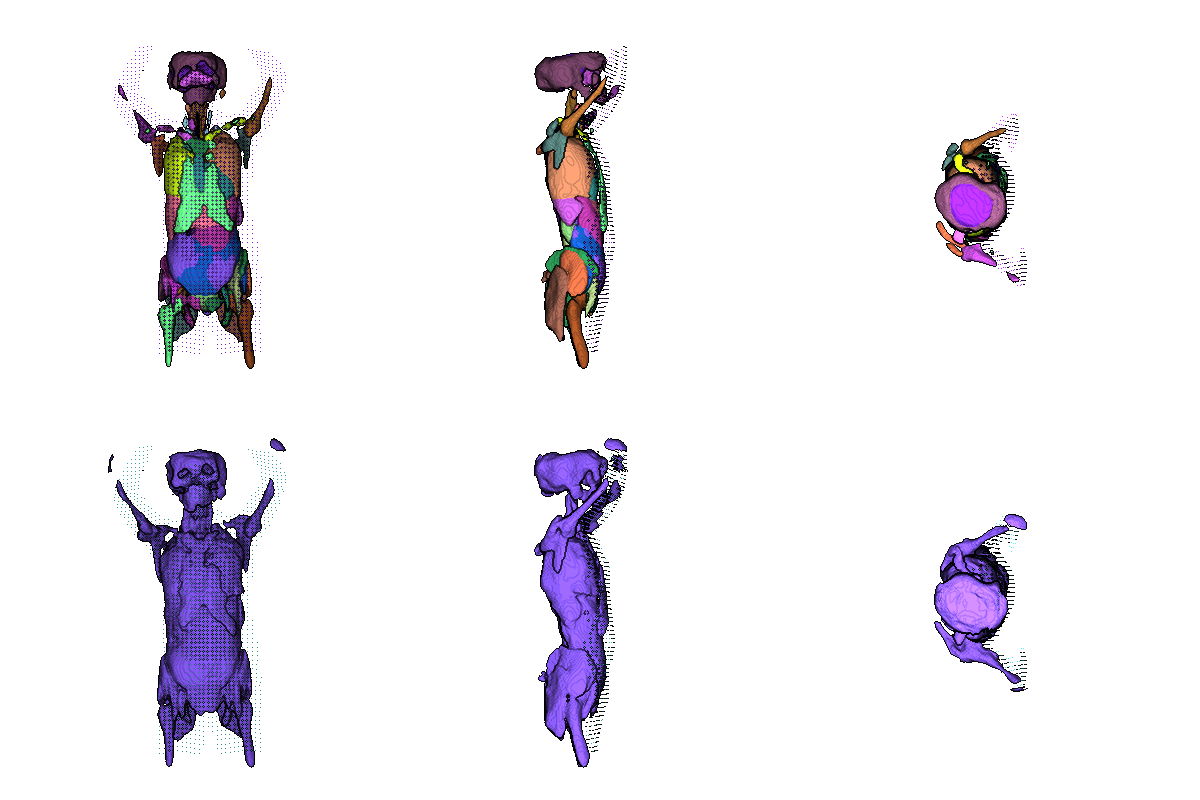}
  \includegraphics[width=0.9880\sfactor, trim={3cm 11cm 23cm 1cm},clip]{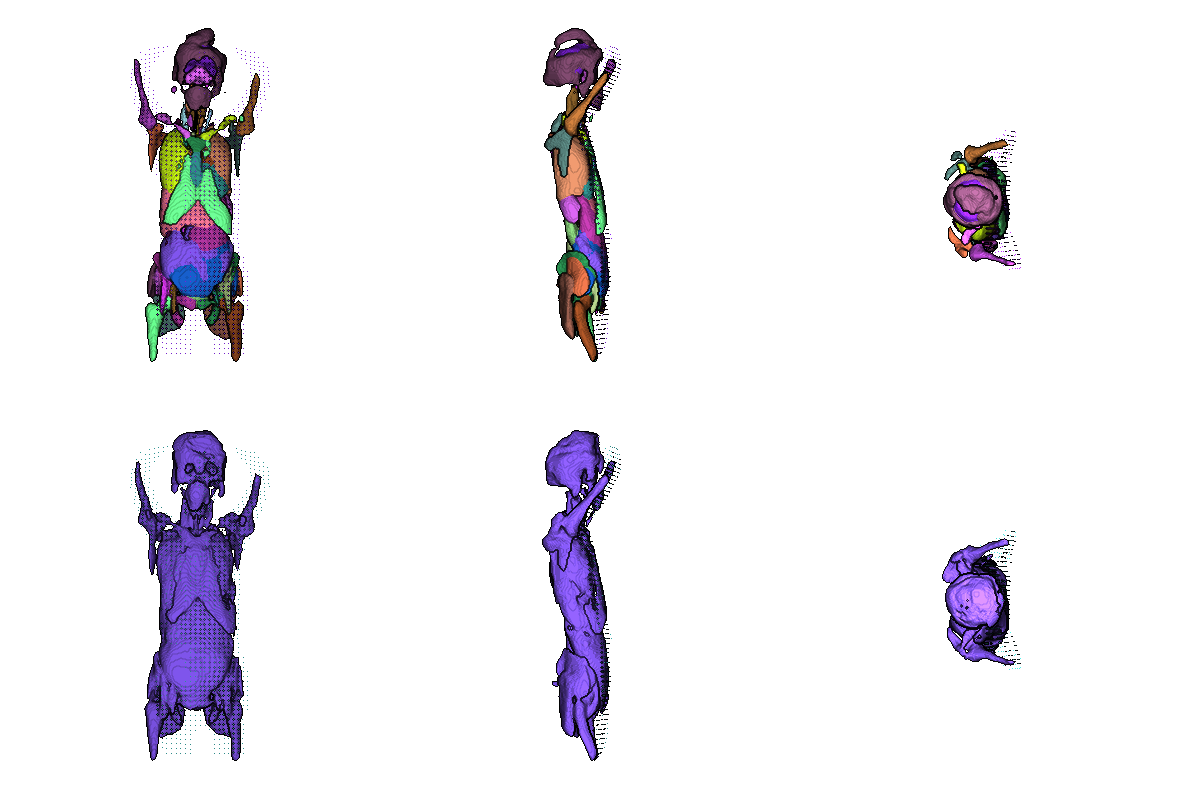}
\caption{Reconstructions from real world data of $12$ individuals ($7$ female, $5$ male). The reconstructions were performed on a point cloud obtained from a single depth image. The depth information was captured using a PMD flexx2 3D camera (pmdtechnologies AG, Germany).}
\label{fig:real_world_data}
\end{figure}

\definecolor{light_orange}{RGB}{253, 182, 100}
\definecolor{dark_orange}{RGB}{203, 122, 0}
\definecolor{very_dark_orange}{RGB}{183, 102, 0}
\definecolor{light_blue}{RGB}{72,153,255}
\definecolor{dark_blue}{RGB}{36,75,128}

\pgfplotsset{
  errorBars/.style={
    error bars/error bar style={
      thick,light_blue
    },
    error bars/y dir=both,
    error bars/y explicit,
  }
}

\begin{figure*}[htbp]
\centering

    \centering
    \small
    
    \begin{tabular}{|c|l|c|l|c|l|c|l|}
    
    \hline
    \textbf{ID} & \textbf{Structure} & \textbf{ID} & \textbf{Structure} & \textbf{ID} & \textbf{Structure} & \textbf{ID} & \textbf{Structure} \\
    \hline
    00 & None & 17 & FemurL & 34 & IliacVeinR & 51 & Sacrum \\
    01 & AdrenalGlandL & 18 & FemurR & 35 & IliopsoasL & 52 & ScapulaL \\
    02 & AdrenalGlandR & 19 & Gallbladder & 36 & IliopsoasR & 53 & ScapulaR \\
    03 & Aorta & 20 & GluteusMaxL & 37 & InferiorVenaCava & 54 & Skull \\
    04 & ArteryBrachiocephalic & 21 & GluteusMaxR & 38 & InternalJugularV.L & 55 & SmallBowel \\
    05 & ArteryInternalCarotidL & 22 & GluteusMedL & 39 & InternalJugularV.R & 56 & Spleen \\
    06 & ArteryInternalCarotidR & 23 & GluteusMedR & 40 & KidneyL & 57 & Sternum \\
    07 & ArterySubclavianL & 24 & GluteusMinL & 41 & KidneyR & 58 & SternumManubrium \\
    08 & ArterySubclavianR & 25 & GluteusMinR & 42 & Liver & 59 & Stomach \\
    09 & Bladder & 26 & Heart & 43 & LungsL & 60 & ThyroidL \\
    10 & Brain & 27 & HipL & 44 & LungsR & 61 & ThyroidR \\
    11 & Cart.CommonCarotidL & 28 & HipR & 45 & NasalCavity & 62 & Trachea \\
    12 & Cart.CommonCarotidR & 29 & HumerusL & 46 & Pancreas & 63 & VeinBrachiocephalicL \\
    13 & ClavicleL & 30 & HumerusR & 47 & Portal\&SplenicVein & 64 & VeinBrachiocephalicR \\
    14 & ClavicleR & 31 & IliacArteryL & 48 & PulmonaryArtery & 65 & VertebraeCervical \\
    15 & Colon & 32 & IliacArteryR & 49 & RibCartilage & 66 & VertebraeLumbar \\
    16 & Esophagus & 33 & IliacVeinL & 50 & Ribs & 67 & VertebraeThoracic \\
    
    \hline
    \end{tabular}
  \label{fig:results_structure_ids}

    \centering
    \begin{tikzpicture}
    \node[anchor=north west] at (14.2,3.8) {{$\downarrow$ Better}};
    \begin{axis}[
        height=5.5cm,
        xticklabel={%
            \pgfmathparse{\tick<10?"0":""}%
            \pgfmathresult%
            \pgfmathprintnumber{\tick}%
        },
        enlarge x limits=0.02,
        ybar,
        bar width=3pt,
        xlabel={ID of Anatomical Structure},
        ylabel={CD [$\text{cm}$]},
        ymin=0,
        ymax=8,
        xtick=data,
        xtick pos=left,
        ymajorgrids=true,
        x tick label style={rotate=90,anchor=east,font=\tiny},
    ]
    \addplot+[errorBars, fill=light_orange, %
        draw=dark_orange, %
        bar width=6pt, %
        bar shift=0,
        error bars/.cd,
        error bar style={color=dark_orange}, %
        y dir=both,
        y explicit,
        ]
    table[
        x=ValueNumber,
        y expr=\thisrow{Mean}*10, %
        y error plus expr=\thisrow{Std}*10, %
        y error minus expr=\thisrow{Std}*10,
        col sep=comma
    ] {data_with_camera_movement/distance_error_estimate.csv};

    \addplot+[errorBars, fill=light_blue, %
        draw=dark_blue, %
        bar width=1.5pt, %
        bar shift=0,
        error bars/.cd,
        error bar style={thin, color=dark_blue}, %
        y dir=both,
        y explicit,
        ]
    table[
        x=ValueNumber,
        y expr=\thisrow{Mean}*10, %
        col sep=comma] {data_baseline_icp_with_camera_moving/distance_error_estimate.csv};
    \end{axis}
    \end{tikzpicture}

    \centering
    \begin{tikzpicture}
    \node[anchor=north west] at (14.2,3.4) {{$\uparrow$ Better}};
    \begin{axis}[
        height=5cm,
        xticklabel={%
            \pgfmathparse{\tick<10?"0":""}%
            \pgfmathresult%
            \pgfmathprintnumber{\tick}%
        },
        enlarge x limits=0.02,
        ybar,
        bar width=3pt,
        xlabel={ID of Anatomical Structure},
        ylabel={IoU},
        ymin=0,
        ymax=0.7,
        ymajorgrids=true,
        xtick=data,
        xtick pos=left,
        x tick label style={rotate=90,anchor=east,font=\tiny},
    ]
    \addplot+[errorBars, fill=light_orange, %
        draw=dark_orange, %
        bar width=6pt, %
        bar shift=0,
        error bars/.cd,
        error bar style={color=dark_orange}, %
        y dir=both,
        y explicit,
        ]
    table[
        x=ValueNumber,
        y=Mean,
        y error plus=Std,
        y error minus=Std,
        col sep=comma] {data_with_camera_movement/iou_estimate.csv};

    \addplot+[errorBars, fill=light_blue, %
        draw=dark_blue, %
        bar width=1.5pt, %
        bar shift=0,
        error bars/.cd,
        error bar style={thin, color=dark_blue}, %
        y dir=both,
        y explicit,
        ]
    table[
        x=ValueNumber,
        y=Mean,
        col sep=comma] {data_baseline_icp_with_camera_moving/iou_estimate.csv};

    \end{axis}
    \end{tikzpicture}

    \centering
    \begin{tikzpicture}
    \node[anchor=south west] at (1.7,3.6) {\tiny{$\leftarrow$~\textcolor{dark_orange}{Occupancy Network $5.1 \pm {3.8}$}, \textcolor{dark_blue}{Baseline $6.4$}}};
    \node[anchor=north west] at (14.2,3.8) {{$\downarrow$ Better}};
    \begin{axis}[
        height=5.5cm,
        xticklabel={%
            \pgfmathparse{\tick<10?"0":""}%
            \pgfmathresult%
            \pgfmathprintnumber{\tick}%
        },
        enlarge x limits=0.02,
        ybar,
        bar width=3pt,
        xlabel={ID of Anatomical Structure},
        ylabel={ESF},
        ymin=1,
        ymax=4.0,
        xtick=data,
        xtick pos=left,
        ytick distance=1, 
        ymajorgrids=true,
        x tick label style={rotate=90,anchor=east,font=\tiny},
    ]
    \addplot+[errorBars, fill=light_orange, %
        draw=dark_orange, %
        bar width=6pt, %
        bar shift=0,
        error bars/.cd,
        error bar style={color=dark_orange}, %
        y dir=both,
        y explicit,
        ]
    table[
        x=ValueNumber,
        y=Mean,
        y error plus=Std,
        y error minus=Std,
        col sep=comma] {data/scale_estimate.csv};

    \addplot+[errorBars, fill=light_blue, %
        draw=dark_blue, %
        bar width=1.5pt, %
        bar shift=0,
        error bars/.cd,
        error bar style={thin, color=dark_blue}, %
        y dir=both,
        y explicit,
        ]
    table[
        x=ValueNumber,
        y=Mean,
        col sep=comma] {data_with_camera_movement/scale_estimate.csv};
    \end{axis}
    \end{tikzpicture}

\caption{
    Table and results of the evaluation on the 50 reserved masks.
    The wider bars (\textcolor{orange}{orange}) are the results from the occupancy network.
    The thinner bars (\textcolor{blue}{blue}) are from the template matching baseline.
    The table shows the ID for each anatomical structure with suffix L (left) and R (right).
    The charts show mean CD, IoU and ESF for each anatomical structure.
    For visual clarity, we only show standard deviations for our occupancy network.
    Larger is better for IoU.
    Smaller is better for CD and ESF.
}
\label{fig:combined}
\end{figure*}

        \subsection{Inference Time} The per-patient occupancy network inference time with hierarchical sampling was $3.2 \pm 0.5$ (mean $\pm$ standard deviation) seconds on a single Nvidia RTX 4090 GPU (Nvidia, Santa Clara, California).
        The inference time for template matching was $1.7 \pm 0.25$ seconds.

        \subsection{Comparison to Baseline}
        Real-world occupancy network estimations for anatomical atlases are qualitatively shown in \Cref{fig:real_world_data}.
        Quantitative results for the baseline and our method are presented in \Cref{fig:combined}.
        In an optimal setting for template matching (with fixed body poses and camera angles), our occupancy network surpasses template matching in the majority of the \gls{as}.
        The occupancy network performs better for $95.5\%$ of \gls{as} considering \gls{cd}.
        For \gls{iou} and \gls{esf} it outperforms the template matching for $68.7\%$ and $82.0\%$ of \glspl{as}, respectively.
        Nevertheless, there are \gls{as} for which our occupancy network performs poorly, such as the \textit{Ribs}.
        We provide the likely reasons in \Cref{sec:results_occupancy_network}.
        The occupancy network and the baseline have an average \gls{cd} of $2.0$~cm and $2.98$~cm across all \gls{as}, respectively.
        
        Unlike the template matching using \gls{icp} for registration, which is inherently rigid, our approach is able to estimate the location of \gls{as} for varying poses, an example is shown in \Cref{fig:overview}.
        When changes in body poses, as described in \Cref{sec:training_data}, are enabled for the evaluation dataset.
        We found the occupancy network to perform similarly on the evaluation data with and without changing poses, with only a reduction of \gls{iou} by an average of $0.01$.

        \subsection{Details on Results of Occupancy Network}
        \label{sec:results_occupancy_network}
        The \gls{iou} of larger structures, such as the \textit{Liver} ($0.5$) or the \textit{Colon} ($0.61$), are higher than of small structures.
        For small structures, such as the \textit{Gallbladder} or the \textit{Thyroid Left and Right}, where small positional offsets (relative to the human size) cause low \gls{iou} values of $0.2$, $0.05$, and $0.1$, respectively.

        Small structures featuring a small \gls{iou} and a high \gls{esf} still exhibit a low \gls{cd}.
        This is because, for small structures, a small increase in \gls{cd} results in large effect on \gls{iou} and \gls{esf}.

        For \textit{Ribs}, poor reconstructions result in performance that is worse than the baseline.
        \textit{Ribs} are complex and space-consuming structures and were only assigned $32$ inside and $32$ outside samples in each training example.
        We suspect that this is an insufficient number of occupancy samples.
        The error in the reconstruction is illustrated in \Cref{fig:real_world_data}, where the ribs are not correctly reconstructed.
        Upon visual inspection, the skull suffers from a similar problem as it is a complex thin and hollow structure.
        Additionally, in many masks used for producing training data, the skull is only partially present.
        This promotes a partial reconstruction, where the top is open.

\section{Limitations}
While our proposed approach demonstrates accurate location estimation for multiple \glspl{as}, several limitations remain:

\textbf{Pose Variation in Training Data:} 
The Atlas Dataset contains individuals with the arms positioned above the head, introducing an inherent pose bias in the training process.
As a result, our model requires that the patient’s arms are similarly positioned during inference.
Outside of applications such as medical \glspl{ct}, a patient's arms may be placed differently.
Increasing the diversity of training poses could mitigate this limitation.
Currently, there are no openly available datasets containing diverse anatomical structures for patients in varying body poses.
\textbf{Reconstruction of Complex Structures:}
The ribs, for instance, are thin, elongated, and numerous.
Our current sampling strategy ($32$ inside and $32$ outside points per \gls{as}) appears insufficient to capture the shape of such complex, space-consuming structures.
This leads to partial or inaccurate reconstructions, as observed in \Cref{fig:real_world_data}.
Adaptive sample densities according to geometric complexity may improve reconstructions.
\textbf{Exclusion of Sex-Specific Organs:}
We do not provide estimates for some internal structures, including all sex-specific organs such as the ovaries or the prostate.
As the used dataset is very limited in size, organs which are frequently missing or error prone were removed.
\textbf{Clothing:}
Our entire training set contains only bare-skin individuals, while real-world patients often wear clothing.
Thick or loose-fitting clothing can obscure the surface shape of the body, thereby limiting our model’s ability to correctly estimate the location of anatomical structures.
Nonetheless, the qualitative results shown in \Cref{fig:real_world_data} are promising, indicating that tight or moderately fitting clothing does not prevent the inference of anatomical atlases.
A more diverse training set, for example using synthetic clothed patients, could improve robustness.
\textbf{Missing Extremities:}
The Atlas Dataset excludes extremities such as the lower legs, feet, and hands.
Therefore, our method cannot localize or reconstruct them.
\textbf{Bounding Box Approximation:}
\gls{aabb}s are not the closest fitting bounding boxes, this can result in a suboptimal \gls{roi} estimate.
For applications that require a more precise estimation, the predicted anatomical atlas should be used instead.
\textbf{Lack of Comparisons:}
Although we surpass the template matching baseline, our method has not been directly compared against how trained medical professionals estimate \gls{as} locations from the outside.
Furthermore, the development of data-driven organ localization methods has only recently been enabled by large segmented datasets, such as Atlas Dataset~\cite{jaus2023towards}.
The current lack of accessible methods able to localize multiple anatomical structures from a single-view observation hinders a comprehensive comparisons.

Despite these limitations, our qualitative results on clothed individuals are promising, suggesting that our approach is robust to moderate deviations between training conditions and real-world data.

\section{Conclusion}
We present a method that estimates the bounding boxes of $67$ \glspl{as} for human bodies using a single low-cost depth sensor.
Applications include the region of interest selection for medical scanners.
We also believe it can act as a valuable redundancy and safety system for downstream automation applications.
We provide a comprehensive list of remaining challenges that should be addressed for applications outside of tomographic imaging.
Trained on augmented virtual data, our system is able to produce an estimated 3D anatomical atlas from a single depth image.
Our approach outperforms localization using template matching on 50 fully segmented masks excluded from the training dataset.
We also provide a real-world qualitative overview of 3D atlases of 12 real-world individuals, both males and females, where the sensor point cloud was captured by a low-cost depth sensor.

{\small
\bibliographystyle{ieee_fullname}
\bibliography{egbib}

\begin{thebibliography}{10}\itemsep=-1pt

\bibitem{annurev-control}
Yuan Bi, Zhongliang Jiang, Felix Duelmer, Dianye Huang, and Nassir Navab.
\newblock Machine learning in robotic ultrasound imaging: Challenges and perspectives.
\newblock {\em Annual Review of Control, Robotics, and Autonomous Systems}, 7(1):null, 2024.

\bibitem{choy20163d}
Christopher~B Choy, Danfei Xu, JunYoung Gwak, Kevin Chen, and Silvio Savarese.
\newblock 3d-r2n2: A unified approach for single and multi-view 3d object reconstruction.
\newblock In {\em Computer Vision--ECCV 2016: 14th European Conference, Amsterdam, The Netherlands, October 11-14, 2016, Proceedings, Part VIII 14}, pages 628--644. Springer, 2016.

\bibitem{comte20233d}
Nicolas Comte, Sergi Pujades, Aur{\'e}lien Courvoisier, Olivier Daniel, Jean-S{\'e}bastien Franco, Fran{\c{c}}ois Faure, and Edmond Boyer.
\newblock 3d inference of the scoliotic spine from depth maps of the back.
\newblock In {\em International Symposium on Computer Methods in Biomechanics and Biomedical Engineering}, pages 159--168. Springer, 2023.

\bibitem{gatidis2022whole}
Sergios Gatidis, Tobias Hepp, Marcel Fr{\"u}h, Christian La~Foug{\`e}re, Konstantin Nikolaou, Christina Pfannenberg, Bernhard Sch{\"o}lkopf, Thomas K{\"u}stner, Clemens Cyran, and Daniel Rubin.
\newblock A whole-body fdg-pet/ct dataset with manually annotated tumor lesions.
\newblock {\em Scientific Data}, 9(1):601, 2022.

\bibitem{girdhar2016learning}
Rohit Girdhar, David~F Fouhey, Mikel Rodriguez, and Abhinav Gupta.
\newblock Learning a predictable and generative vector representation for objects.
\newblock In {\em Computer Vision--ECCV 2016: 14th European Conference, Amsterdam, The Netherlands, October 11-14, 2016, Proceedings, Part VI 14}, pages 484--499. Springer, 2016.

\bibitem{grundy2004obesity}
Scott~M Grundy.
\newblock Obesity, metabolic syndrome, and cardiovascular disease.
\newblock {\em The Journal of Clinical Endocrinology \& Metabolism}, 89(6):2595--2600, 2004.

\bibitem{henrich2024registered}
Pit Henrich, Bal{\'a}zs Gyenes, Paul~Maria Scheikl, Gerhard Neumann, and Franziska Mathis-Ullrich.
\newblock Registered and segmented deformable object reconstruction from a single view point cloud.
\newblock In {\em Proceedings of the IEEE/CVF Winter Conference on Applications of Computer Vision}, pages 3129--3138, 2024.

\bibitem{isensee2021nnu}
Fabian Isensee, Paul~F Jaeger, Simon~AA Kohl, Jens Petersen, and Klaus~H Maier-Hein.
\newblock nnu-net: a self-configuring method for deep learning-based biomedical image segmentation.
\newblock {\em Nature methods}, 18(2):203--211, 2021.

\bibitem{jaus2023towards}
Alexander Jaus, Constantin Seibold, Kelsey Hermann, Alexandra Walter, Kristina Giske, Johannes Haubold, Jens Kleesiek, and Rainer Stiefelhagen.
\newblock Towards unifying anatomy segmentation: Automated generation of a full-body ct dataset via knowledge aggregation and anatomical guidelines.
\newblock {\em arXiv preprint arXiv:2307.13375}, 2023.

\bibitem{jiang2023robotic}
Zhongliang Jiang, Septimiu~E Salcudean, and Nassir Navab.
\newblock Robotic ultrasound imaging: State-of-the-art and future perspectives.
\newblock {\em Medical image analysis}, page 102878, 2023.

\bibitem{Keller_2024_CVPR}
Marilyn Keller, Vaibhav Arora, Abdelmouttaleb Dakri, Shivam Chandhok, J\"urgen Machann, Andreas Fritsche, Michael~J. Black, and Sergi Pujades.
\newblock Hit: Estimating internal human implicit tissues from the body surface.
\newblock In {\em Proceedings of the IEEE/CVF Conference on Computer Vision and Pattern Recognition (CVPR)}, pages 3480--3490, June 2024.

\bibitem{Keller:CVPR:2022}
Marilyn Keller, Silvia Zuffi, Michael~J. Black, and Sergi Pujades.
\newblock {OSSO}: Obtaining skeletal shape from outside.
\newblock In {\em Proceedings IEEE/CVF Conf.~on Computer Vision and Pattern Recognition (CVPR)}, pages 20492--20501, June 2022.

\bibitem{lamb2022deepjoin}
Nikolas Lamb, Sean Banerjee, and Natasha~Kholgade Banerjee.
\newblock Deepjoin: Learning a joint occupancy, signed distance, and normal field function for shape repair.
\newblock {\em ACM Transactions on Graphics (TOG)}, 41(6):1--10, 2022.

\bibitem{SMPL:2015}
Matthew Loper, Naureen Mahmood, Javier Romero, Gerard Pons-Moll, and Michael~J. Black.
\newblock {SMPL}: A skinned multi-person linear model.
\newblock {\em ACM Trans. Graphics (Proc. SIGGRAPH Asia)}, 34(6):248:1--248:16, Oct. 2015.

\bibitem{lorensen1998marching}
William~E Lorensen and Harvey~E Cline.
\newblock Marching cubes: A high resolution 3d surface construction algorithm.
\newblock In {\em Seminal graphics: pioneering efforts that shaped the field}, pages 347--353. 1998.

\bibitem{mescheder2019occupancy}
Lars Mescheder, Michael Oechsle, Michael Niemeyer, Sebastian Nowozin, and Andreas Geiger.
\newblock Occupancy networks: Learning 3d reconstruction in function space.
\newblock In {\em Proceedings of the IEEE/CVF conference on computer vision and pattern recognition}, pages 4460--4470, 2019.

\bibitem{modrzejewski2018soft}
Richard Modrzejewski, Toby Collins, Adrien Bartoli, Alexandre Hostettler, and Jacques Marescaux.
\newblock Soft-body registration of pre-operative 3d models to intra-operative rgbd partial body scans.
\newblock In {\em Medical Image Computing and Computer Assisted Intervention--MICCAI 2018: 21st International Conference, Granada, Spain, September 16-20, 2018, Proceedings, Part IV 11}, pages 39--46. Springer, 2018.

\bibitem{STAR:2020}
Ahmed A~A Osman, Timo Bolkart, and Michael~J. Black.
\newblock {STAR}: A sparse trained articulated human body regressor.
\newblock In {\em European Conference on Computer Vision (ECCV)}, pages 598--613, 2020.

\bibitem{park2019deepsdf}
Jeong~Joon Park, Peter Florence, Julian Straub, Richard Newcombe, and Steven Lovegrove.
\newblock Deepsdf: Learning continuous signed distance functions for shape representation.
\newblock In {\em Proceedings of the IEEE/CVF conference on computer vision and pattern recognition}, pages 165--174, 2019.

\bibitem{SMPL-X:2019}
Georgios Pavlakos, Vasileios Choutas, Nima Ghorbani, Timo Bolkart, Ahmed A.~A. Osman, Dimitrios Tzionas, and Michael~J. Black.
\newblock Expressive body capture: {3D} hands, face, and body from a single image.
\newblock In {\em Proceedings IEEE Conf. on Computer Vision and Pattern Recognition (CVPR)}, pages 10975--10985, 2019.

\bibitem{qi2017pointnet++}
Charles~Ruizhongtai Qi, Li Yi, Hao Su, and Leonidas~J Guibas.
\newblock Pointnet++: Deep hierarchical feature learning on point sets in a metric space.
\newblock {\em Advances in neural information processing systems}, 30, 2017.

\bibitem{Moradi2022}
Septimiu~E. Salcudean, Hamid Moradi, David~G. Black, and Nassir Navab.
\newblock Robot-assisted medical imaging: A review.
\newblock {\em Proceedings of the IEEE}, 110(7):951--967, 2022.

\bibitem{SHETTY2023107383}
Karthik Shetty, Annette Birkhold, Srikrishna Jaganathan, Norbert Strobel, Bernhard Egger, Markus Kowarschik, and Andreas Maier.
\newblock Boss: Bones, organs and skin shape model.
\newblock {\em Computers in Biology and Medicine}, 165:107383, 2023.

\end{thebibliography}
}

\end{document}